\documentclass[11pt]{article}

\usepackage[final]{acl}
\usepackage{float}
\usepackage{times}
\usepackage{latexsym}
\usepackage{amsmath}
\usepackage{amssymb}
\usepackage[T1]{fontenc}
\usepackage{placeins}
\usepackage[utf8]{inputenc}

\usepackage{microtype}
\usepackage{tabularx}
\usepackage{inconsolata}

\usepackage{graphicx}
\usepackage{url}
\usepackage{multirow}
\usepackage{booktabs} 
\usepackage{tabularx}
\usepackage[normalem]{ulem}
\usepackage{subcaption}

\title{Towards Diagnostic Feedback for Related Work Writing: \\A Span-Level Benchmark}

\author{Iman Al Sikaiti \\
  MBZUAI, UAE \\
  \texttt{iman.alsikaiti@mbzuai.ac.ae} \\\And
  Ekaterina Kochmar \\
  University of Victoria \\
  \texttt{ekaterinakochmar@uvic.ca} \\}

\begin{document}
\maketitle
\begin{abstract}
The Related Work section is central to academic writing, yet prior work has largely focused on generation and citation classification rather than the analysis of writing issues within the text. To support model evaluation in this setting, we formulate span-level error detection for Related Work sections and introduce a taxonomy of four categories: \textit{Format}, \textit{Unsupported Claim}, \textit{Coherence}, and \textit{Lacks Synthesis}. We construct an annotated dataset from {\tt NLPeer} with 137 documents and 395 spans, and generate 1,100 synthetic training spans using {\tt GPT-5}. Using this dataset, we benchmark prompting-based baselines and encoder models for detecting writing issue spans, alongside a citation-aware post-processing pipeline for short \textit{Unsupported Claim} spans. Encoder models outperform prompting approaches in span localization, while prompting methods tend to identify broad regions without precise boundaries. We also find that synthetic spans match real spans structurally but show reduced lexical diversity, affecting boundary alignment. Our work provides a benchmark dataset and empirical analysis for automated evaluation of Related Work writing quality, which we make publicly available.\footnote{The data and code used in this study can be found at \href{https://github.com/AerdnaNami/Towards-Diagnostic-Feedback-for-Related-Work-Writing}{github.com/AerdnaNami/Towards-Diagnostic-Feedback}}
\end{abstract}

\section{Introduction}
\label{sec:introduction}

The Related Work Section (RWS) is a key component of an academic paper. It demonstrates an author's familiarity with a field, identifies research gaps, sets up the argumentative context of the paper, and positions the paper's contributions within existing research \cite{cmu_lit_review, teufel-etal-2006-automatic}. In doing so, it not only provides background information, but also helps the reader understand how the proposed work relates to and advances prior efforts.

Despite its importance, relatively little work has focused on identifying fine-grained writing issues within RWSs. Existing work has primarily focused on generating entire Related Work sections \cite{sahinuc2026expertpreferencebasedevaluationautomated, li-ouyang-2024-related} or classifying citation intent \cite{jurgens-etal-2018-measuring, lauscher-etal-2022-multicite}, rather than detecting where specific writing problems occur in the text. As a result, there is no established task formulation or benchmark for localizing issues such as weak narrative structure, missing attribution, lack of synthesis, or citation formatting errors.

\begin{figure}[t]
    \centering
    \includegraphics[width=\linewidth]{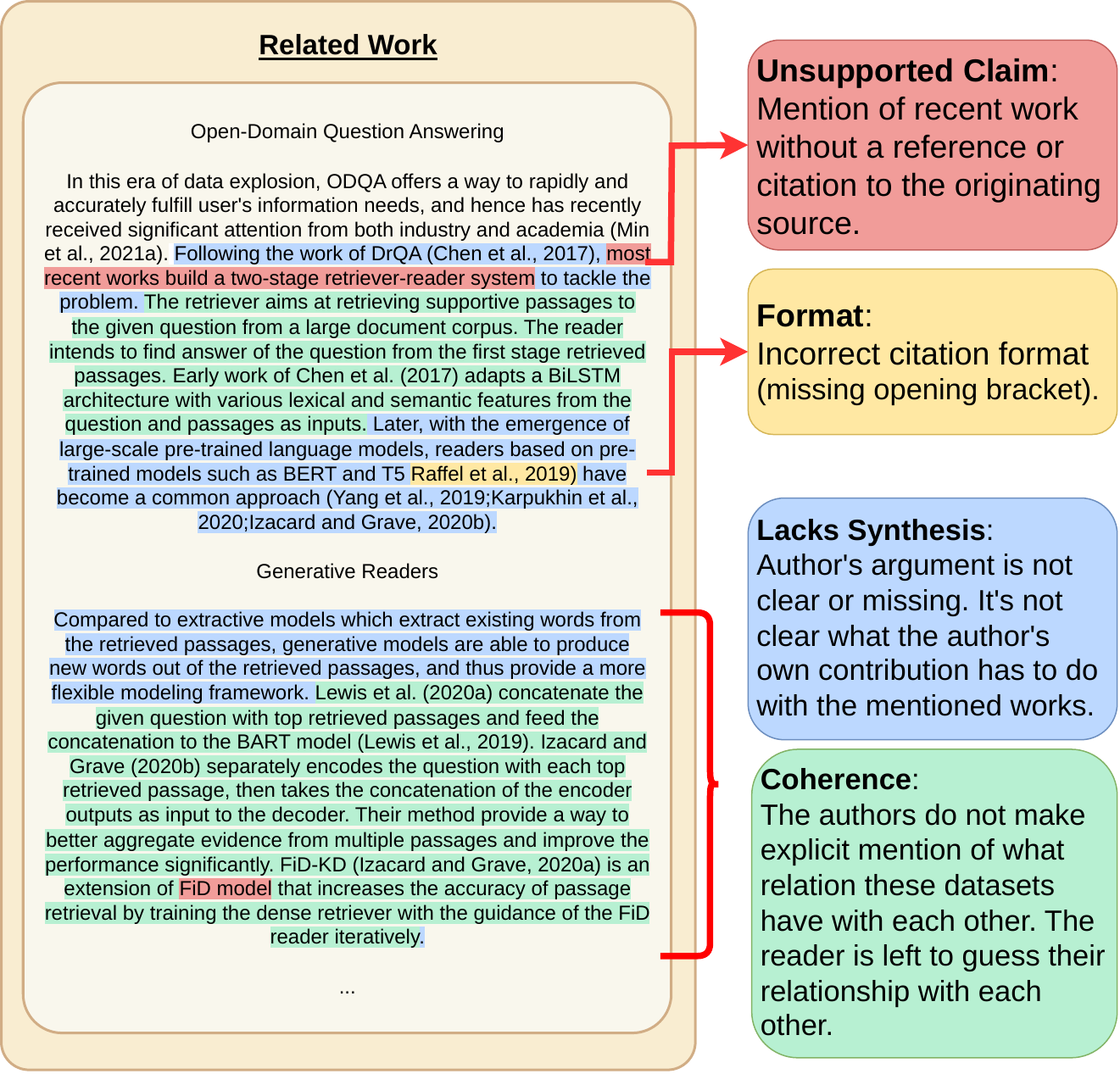}
    \caption{Examples of span-level writing issues identified in Related Work sections in our data.}
    \label{fig:error_examples_intro}
\end{figure}

Prior work on Related Work generation highlights several recurring writing challenges. \citet{sahinuc2026expertpreferencebasedevaluationautomated} and \citet{li-ouyang-2024-related} show that LLM-generated RWSs often fail to adequately emphasize citations, leading to vague and unfocused writing. Such text may summarize prior work without clearly articulating the author's own perspective or positioning the current contribution. To capture this issue, we define a \textbf{Lacks Synthesis} category for text regions that fail to express an overarching argument or narrative, though we note that boundaries for such paragraph- or discourse-level issues are inherently approximate (see Figure \ref{fig:error_examples_intro}).

Related Work sections also require clear connections between cited works. \citet{li-ouyang-2024-related} report that generated RWSs often do not draw sufficient connections between cited papers, producing text that resembles a concatenation of summaries rather than a cohesive argument. Similarly, \citet{li-ouyang-2025-explaining} report a human preference for text that includes transition sentences that relate one cited work to another. Motivated by this, we define a \textbf{Coherence} category for text regions where relationships between cited works are ambiguous or not clearly expressed. Like \textbf{Lacks Synthesis}, this is a discourse-level category whose span boundaries are inherently approximate.

A further challenge concerns attribution. Prior work shows that references to earlier work are often expressed implicitly rather than through explicit citations, making it difficult to determine whose work or idea is being discussed \cite{siddharthan-teufel-2007-whose}. This can result in claims that lack clear evidence. We therefore define an \textbf{Unsupported Claim} category for statements that imply reference to prior work but fail to provide explicit attribution.

Finally, writing quality also depends on the adherence to citation conventions. Citation formatting errors, such as missing brackets, inconsistent citation styles, or incomplete references, can affect readability and clarity. We include a \textbf{Format} category to capture these surface-level but practically important issues.

These issues make Related Work feedback a challenging setting for automated evaluation. Useful feedback must localize the problematic region of text, but many of the relevant issues also depend on discourse-level and argumentative reasoning. For example, identifying a missing citation can require recognizing an unsupported claim, while identifying weak argument synthesis requires understanding how cited works relate to the author’s contribution. Because the boundaries of discourse-level issues can be subjective, we use span-level F1 and Cohen's kappa to separately evaluate the existence versus the extent of issues: F1 captures whether an issue is identified at all, while Cohen's kappa accounts for chance agreement across raters when judging how far an issue spans. We therefore formulate Related Work evaluation as {\em a span-level diagnostic task designed to study the interaction between localization and argument-level reasoning}.

Rather than proposing a new model architecture, this work introduces a benchmark for studying this problem. We compare prompting-based LLMs, which may capture broad discourse cues, against encoder models, which are better suited to span localization, and analyze where each approach stands.
We make the following contributions:
\begin{itemize}
\setlength{\itemsep}{0pt}
    \item We introduce a taxonomy of writing-related errors in Related Work and Introduction sections,\footnote{For this task, we treat Introduction as similar to the Related Work section if it introduces the background for the work presented in the paper.} covering citation formatting, unsupported claims, coherence, and synthesis.
    \item We construct an annotated dataset of 137 real Introduction and Related Work sections consisting of 395 annotated spans from {\tt NLPeer} \cite{dycke-etal-2023-nlpeer}, along with 1,100 synthetic training spans grounded in real documents.
    \item We formulate the task as span-level error detection and evaluate prompting-based baselines against encoder-based models.
    \item We analyze the factors that affect performance, including citation-aware post-processing, {\tt SciERC} \cite{luan-etal-2018-multi} pretraining, and the structural and lexical quality of synthetic data.
    \item We distinguish between categories that allow for precise localization (e.g. Format, Unsupported Claim) and those requiring approximate discourse-level judgments (Coherence, Lacks Synthesis), and show that Pairwise F1 and Cohen's kappa capture complementary aspects of performance on each.
\end{itemize}

\section{Related Work}
\label{sec:rel_work}

\subsection{Evaluation of Related Work Generation}
Early studies in Related Work generation emphasize the importance of including informative and indicative sentences \cite{kan2001informative,Hoang2010TowardsAR}, where informative sentences provide details of prior work, while indicative sentences establish relationships between cited works, helping to create a coherent narrative about the author's contribution. Despite that, early approaches to Related Work generation \citep{Hoang2010TowardsAR, chen-automatic-gen-related-works, Yasunaga_Kasai_Zhang_Fabbri_Li_Friedman_Radev_2019} often framed the task as extractive multi-document summarization that struggled to capture the indicative structure found in effective Related Work sections. More recent approaches adopt abstractive methods to generate Related Work sections using large language models (LLMs) \cite{chen-etal-2021-capturing, sahinuc2026expertpreferencebasedevaluationautomated} or sequence-to-sequence models \cite{aburaed_automatic_2020}, which improve the analysis of indicative structure. However, these improvements remain insufficient to consistently produce a coherent narrative \cite{li-ouyang-2024-related}.

While this line of work highlights the importance of indicative structure, it does not provide mechanisms for localizing where indicative relations are missing, weak, or unclear within the text. In contrast, we formulate a span-level benchmark for identifying regions of Related Work writing where the narrative function breaks down, such as missing connections between cited works or insufficient synthesis of prior work with the author's contribution.

\begin{table*}[t]
\centering
\small
\setlength{\tabcolsep}{4pt}
\renewcommand{\arraystretch}{1.1}

\begin{tabularx}{\textwidth}{
p{0.15\textwidth}
p{0.20\textwidth}
X
p{0.18\textwidth}
}
\toprule
\textbf{Category} 
& \textbf{Writing function} 
& \textbf{Definition} 
& \textbf{Span level} \\
\midrule

\textit{Format} 
& Citation convention 
& Surface-level citation formatting errors, such as missing brackets, incorrect parenthetical or narrative citation style, or incomplete references. 
& Citation-level \\

\textit{Unsupported Claim} 
& Evidential grounding 
& Claims or mentions of prior work that lack explicit citation support. Short cases usually involve uncited datasets, models, methods, or tasks; long cases involve broader argumentative claims without evidence. 
& Phrase or sentence \\

\textit{Coherence} 
& Relation between cited works 
& Missing or unclear transitions between cited works, leaving their relationship implicit or ambiguous. 
& Multi-sentence or paragraph \\

\textit{Lacks Synthesis} 
& Authorial positioning 
& Prior work is summarized without connecting it to the author's own argument, research gap, or contribution. 
& Paragraph or subsection \\

\bottomrule
\end{tabularx}

\caption{Taxonomy of span-level writing issues in Related Work sections.}
\label{tab:error_taxonomy}
\end{table*}

\subsection{Analysis of Related Work Quality}
Evaluation frameworks increasingly assess whether RWS text captures indicative relationships between cited works rather than merely summarizing them. One such work is \citet{li-ouyang-2025-explaining}, who show that generated Related Work sections often lack clear connections between cited works and fail to construct a cohesive argument. Similarly, \citet{sahinuc2026expertpreferencebasedevaluationautomated} introduce positioning-based evaluation aspects, such as \textit{Positioning Existence} and \textit{Positioning Type}, emphasizing the role of indicative structure in persuasive RWSs. Their findings show that even strong LLMs, such as {\tt o3-mini} \cite{openai2025o3mini}, often fail to coherently cite and position prior work. This line of work is further extended by \citet{sahinuc2026rewardmodelingscientificwriting}, who show that scientific writing evaluation requires domain knowledge, reasoning, and explicit criteria.

While these studies provide important insights into high-quality writing, they primarily evaluate LLM-generated text rather than human-authored Related Work sections. In contrast, we define common writing errors in human-written RWSs and detect them at the span level, enabling fine-grained analysis of writing quality to support revision.

\subsection{Span-level Error Localization}

Our work is also related to span-level error localization in other NLP evaluation settings. In machine translation evaluation, MQM-based approaches evaluate translation quality by identifying localized errors and assigning categories such as accuracy, fluency, terminology, and severity. For example, \citet{kocmi-federmann-2023-gemba-mqm} introduce GEMBA-MQM, a GPT-based metric that identifies translation quality error spans and assigns MQM error categories. Similarly, \citet{guerreiro-etal-2024-xcomet} propose xCOMET, an open-source learned metric that combines sentence-level quality estimation with fine-grained error span detection.

Similar to this line of work, we formulate quality evaluation as the identification of localized error spans rather than assigning only holistic scores. However, our setting differs in both domain and reasoning requirements. Instead of evaluating translation quality against a source segment, we evaluate Related Work writing through citation use, attribution, discourse coherence, and synthesis of prior work. We therefore treat MQM-style methods as related work on span-level quality evaluation rather than direct baselines, since our task evaluates text-internal writing issues in Related Work sections rather than source-conditioned translation errors.

\section{Methodology}
\label{sec:method}

\subsection{Error Categories}
\label{sec:error_cat}
We define four error categories based on prior work on effective RWS writing \cite{cmu_lit_review}, as well as recurring writing-related issues found in the {\tt Expos{\'i}a} \cite{zyska2026exposiateachingassessmentacademic} and {\tt NLPeer} \cite{dycke-etal-2023-nlpeer} datasets. The categories capture complementary dimensions of Related Work quality: citation conventions, evidential grounding, coherence between cited works, and synthesis or positioning of the author's contribution. Table~\ref{tab:error_taxonomy} summarizes the taxonomy with examples presented in Table~\ref{tab:category_examples}. 

We select these categories according to three criteria. First, the issue must be observable from the Related Work or Introduction section itself, without requiring external verification against the cited papers. Second, the issue must be localizable as a span or feedback region in the text. Third, the issue must correspond to a known rhetorical function of Related Work writing: following citation conventions, grounding claims in prior work, relating cited works to one another, or positioning the author's contribution.

This taxonomy is not intended to exhaustively cover all possible Related Work issues. Instead, it focuses on writing issues that can be identified from the RWS text itself, rather than issues requiring external verification against the cited literature, such as factual misrepresentation, missing coverage of relevant work, or biased citation selection. For discourse-level categories such as \textit{Coherence} and \textit{Lacks Synthesis}, the annotated spans should be interpreted as approximate feedback regions indicating where the issue becomes salient, rather than as precise token-level error boundaries.

\subsection{Dataset}
\label{sec:dataset}
\paragraph{Gold Dataset} We construct the gold dataset from the ARR subset of {\tt NLPeer} \cite{dycke-etal-2023-nlpeer}, which contains 477 papers. To identify papers with relevant writing issues, we filter associated reviewer comments using {\tt GPT-5} \cite{OpenAI2025GPT5}, as such comments often refer implicitly to issues in the Introduction or Related Work sections. This filtering step yields 117 papers. We then extract the corresponding Introduction and Related Work sections from TEI files and remove noisy or irregular text, resulting in 137 documents for annotation.

Four annotators, all fluent in English and with at least a bachelor's degree in Computer Science or a related field, annotated the data using predefined guidelines (see Appendix~\ref{app:annotation_guidelines}). A pilot round on 10 documents was used to refine the guidelines, followed by annotation of 10 overlapping documents for agreement estimation. Three annotators were additionally assigned 39 unique documents each, while the remaining annotator was assigned 19 unique documents. Final spans for the overlapping documents were determined through majority voting and discussion.

The final evaluation dataset contains 395 span-level annotations across 137 unique
documents: 96 \textit{Format} spans, 189 \textit{Unsupported Claim} spans, 42 \textit{Coherence} spans, and 68 \textit{Lacks Synthesis} spans.

\paragraph{Inter-Annotator Agreement}
Annotation consistency is measured using mean pairwise Cohen's $\kappa$, Fleiss' $\kappa$, and Pairwise F1 \cite{hripcsak_agreement_2005, deleger_building_2012}, as reported in Table~\ref{tab:ann_agreements}. We report these metrics together because they capture complementary aspects of agreement: $\kappa$-based metrics measure chance-corrected token-level consistency, while Pairwise F1 captures span overlap and is less sensitive to small boundary differences. For agreement calculations, annotations in the overlapping documents are converted into category-specific binary token sequences, where tokens inside an annotated span are marked positive and all other tokens are marked negative. Mean pairwise Cohen's $\kappa$ is computed by averaging Cohen's $\kappa$ across annotator pairs, while Fleiss' $\kappa$ is computed over the same token-level annotation units across annotators. Pairwise F1 is used as an overlap-based span agreement measure.

\begin{table*}[t]
\centering
\large
\begin{tabular}{lcccc}
\toprule
\textbf{Category} 
& \textbf{Cohen’s $\kappa$} 
& \textbf{Fleiss’ $\kappa$} 
& \textbf{Pairwise F1} \\
\midrule
Unsupported Claim & 0.40 & 0.40 & 0.33 \\
Format            & 0.83 & 0.89 & 0.49 \\
Coherence         & 0.33 & 0.32 & 0.69 \\
Lacks Synthesis   & 0.43 & 0.22 & 0.37 \\
\bottomrule
\end{tabular}
\caption{Inter-annotator agreement across error categories.}
\label{tab:ann_agreements}
\end{table*}

\paragraph{Synthetic Training Dataset} 
Due to the limited size of the gold-standard dataset, we use the remaining documents from the {\tt NLPeer} ARR subset to generate synthetic spans that are only used for training. We prompt {\tt GPT-5} to insert category-specific errors into real documents, using the prompt in Appendix~\ref{sec:appx_prompts}. Generation is constrained by empirically sampling span lengths and span counts from the gold-standard annotations.

Generated candidates are then verified by a critic LLM, again using {\tt GPT-5}, as an automatic quality-control step to check whether each span follows the intended category definition and boundary constraints. Using the same model keeps the generation and filtering stages consistent in their interpretation of the prompt and output format. However, we do not treat this step as independent validation, since the critic may share biases with the generator. 

To further characterize the quality of the critic-filtered data, we conduct a manual audit of a stratified sample of synthetic spans. Specifically, we have manually audited 100 synthetic spans, stratified evenly across the four categories, for category validity and boundary plausibility. Overall, we find that 80\% of the audited spans satisfy both criteria. Validity is highest for \textit{Unsupported Claim} (88\%), followed by \textit{Format} and \textit{Lacks Synthesis} (84\% each), and lowest for \textit{Coherence} (64\%). This suggests that the synthetic training dataset is useful for training, but contains non-trivial noise, such as spans that only partially satisfy the intended category definition, fail to meet all category-specific requirements, or have less plausible boundaries. This issue is especially pronounced for discourse-level categories, where the distinction between weak coherence and weak synthesis can be more ambiguous. Further details are provided in Table~\ref{tab:silver_audit}.

\subsection{Task Setup \& Training}
The task is formulated as span-level error detection. Given an input document \(X\) from a Related Work or Introduction section, the goal is to identify contiguous spans that correspond to one of four error categories: \textit{Unsupported Claim}, \textit{Format}, \textit{Coherence}, and \textit{Lacks Synthesis}. Unlike prior work that evaluates writing quality with document-level scores, this formulation localizes writing issues directly in the text.

Since error categories occur at different levels of granularity, the modeling setup is category-specific. \textit{Unsupported Claim} and \textit{Format} are treated as sentence-level BIO tagging tasks, where contiguous \texttt{B}/\texttt{I} predictions are converted into spans. Models are trained separately for each category using class-weighted loss to address the imbalance between the dominant \texttt{O} label and span labels. In contrast, \textit{Coherence} and \textit{Lacks Synthesis} are modeled as paragraph-level binary classification tasks over overlapping text windows. Windows predicted as positive are mapped back to their original document offsets, and overlapping positive windows are merged to form final predicted spans. The selected window size, stride, coverage threshold, and BIO label weights are reported in Appendix~\ref{app:hyperparameters}.

For short \textit{Unsupported Claim} spans, a different setup is required because the span itself is short, but the decision depends on wider document and citation context. These spans often correspond to mentions of datasets, models, methods, or scientific terms, making scientific span extraction useful for identifying candidate mentions. However, unlike a standard entity recognition task, a mention should only be treated as unsupported if it is a first mention and is not accompanied by citation support. This requires checking the mention's position within the document and its surrounding citation context, since later mentions may be valid if the work has already been introduced earlier. Encoder models are therefore first adapted to scientific span extraction using {\tt SciERC} pretraining \cite{luan-etal-2018-multi}. Token predictions are converted into candidate spans using a binary threshold, with isolated predictions filtered by a separate confidence threshold. A citation-aware post-processing step then removes spans that are followed by citation support or correspond to non-initial mentions. This makes short \textit{Unsupported Claim} detection distinct from purely local categories such as \textit{Format}, as well as from broader discourse-level categories such as \textit{Coherence} and \textit{Lacks Synthesis}.

\begin{figure*}[!htbp]
    \centering
    \includegraphics[width=0.9\linewidth]{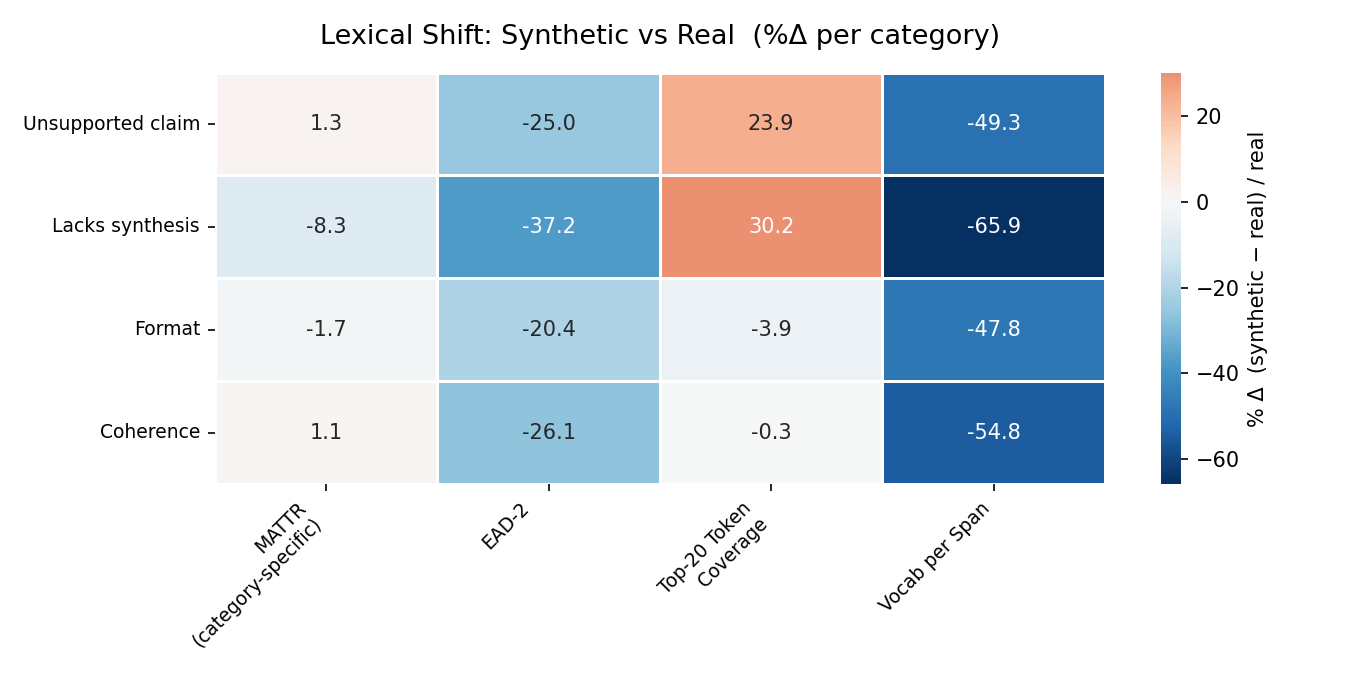}
    \caption{Relative distributional shift (synthetic vs. real) across error categories for selected metrics, including MATTR, EAD-2, Top-20 token coverage, and Vocabulary per span. Values near zero indicate close alignment between synthetic and real samples, while larger deviations reflect distributional divergence.}
    \label{fig:heatmap_relative_shift}
\end{figure*}

\section{Experiments \& Evaluation}
\label{sec:exp_eval}

\subsection{Models Compared}
\paragraph{Baseline} We use zero-shot and few-shot prompting with {\tt GPT-5} \cite{OpenAI2025GPT5} and compare its performance against encoder-based models trained on the synthetic training dataset. The baseline prompts include the category definition, length constraints (whether a category is limited to a phrase, sentence, or paragraph level), and positive and negative examples with few-shot prompting. We use GPT-5 prompting as a strong closed-model baseline for testing whether general-purpose LLMs can identify writing-issue spans without task-specific training. Since the goal of this work is to benchmark span-level Related Work feedback rather than optimize prompting strategies, this baseline serves as a reference point for comparing unsupervised LLM behavior against supervised encoder-based span localization. More extensive comparisons across prompting strategies and open-weight LLMs are left for future work.

\paragraph{Encoder-based Models} The encoder models that we fine-tune for this task are {\tt XLM-RoBERTa} \cite{xlm-roberta}, {\tt SciBERT} \cite{scibert}, {\tt ModernBERT} \cite{modernbert}, and {\tt NeoBERT} \cite{neobert}. We chose to utilize encoder models for this task due to their established capabilities in span extraction that surpass decoder models' capabilities, since decoder models tend to over-generate and change the format of the input text \cite{semin2026strategiesspanlabelinglarge, jarolim2025llmsextracthumanlikefinegrained}. Furthermore, we chose to train each model on a single category, to introduce as little noise from other categories as possible. The encoder models are selected to cover scientific-domain pretraining ({\tt SciBERT}), general multilingual encoding ({\tt XLM-R}), and recent bidirectional encoder architectures ({\tt ModernBERT}, {\tt NeoBERT}).

As an additional experiment with encoder models, we evaluate the effect of domain- and task-specific pretraining by incorporating {\tt SciERC} dataset \cite{luan-etal-2018-multi} for \textit{Short Unsupported Claim} detection. This allows us to analyze whether prior training to detect scientific entity and span extraction improves performance on this subcategory.

\begin{figure*}[t]
    \centering

    \begin{subfigure}[t]{0.47\textwidth}
        \centering
        \includegraphics[width=\linewidth]{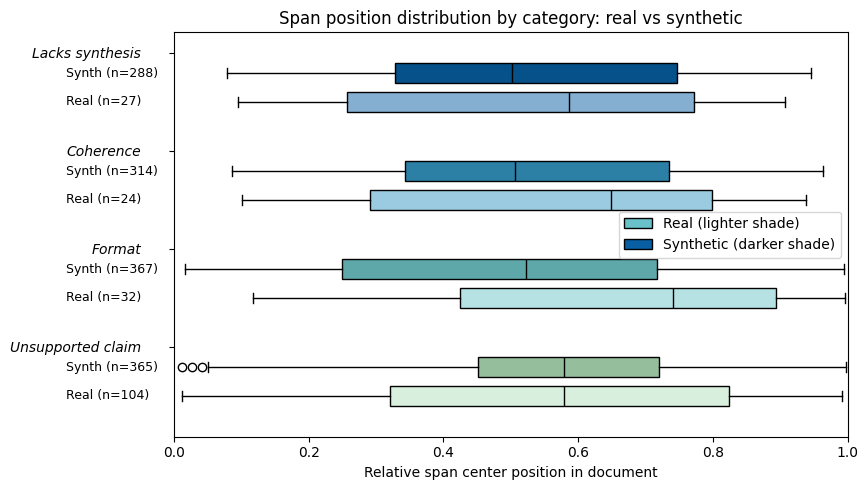}
        \caption{Span position distribution.}
        \label{fig:center_pos_comparison}
    \end{subfigure}
    \hfill
    \begin{subfigure}[t]{0.47\textwidth}
        \centering
        \includegraphics[width=\linewidth]{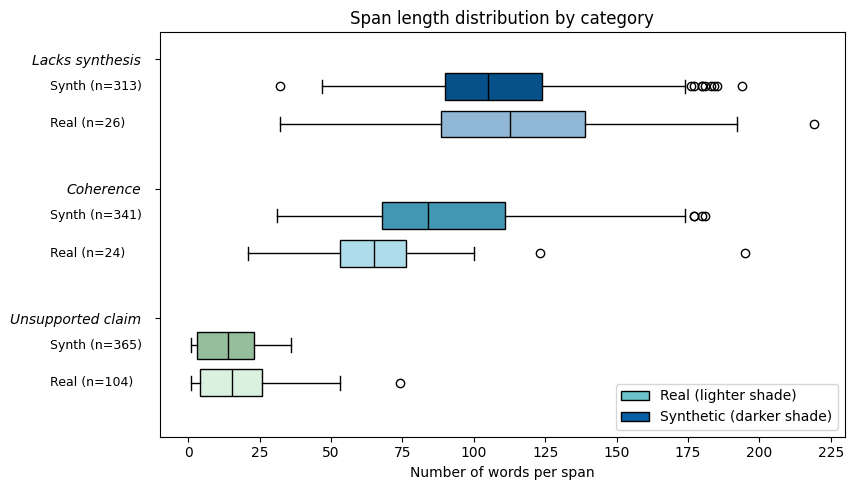}
        \caption{Span length distribution excluding Format.}
        \label{fig:span_length_non_format}
    \end{subfigure}

    \vspace{0.75em}

    \begin{subfigure}[t]{0.47\textwidth}
        \centering
        \includegraphics[width=\linewidth]{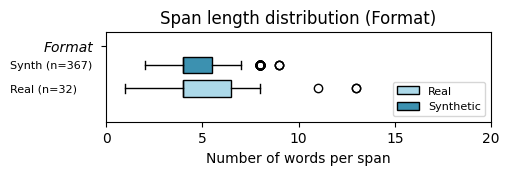}
        \caption{Span length distribution for Format.}
        \label{fig:format_span_length}
    \end{subfigure}

    \caption{Structural comparison of synthetic and real spans. (a) shows span center positions across documents, where 0 indicates the start of the text and 1 indicates the end. (b) shows span length distributions for non-Format categories. (c) shows Format separately because its span lengths have a substantially smaller value range than the other categories.}
    \label{fig:synth_vs_real_structural}
\end{figure*}

\subsection{Experimental Setting}
We use the synthetic training dataset and the gold-standard dataset defined in Section~\ref{sec:dataset}, for training purposes and evaluation, respectively. 

Encoder models are fine-tuned using a learning rate of $3 \times 10^{-5}$ and a batch size of 16, and trained for 3 epochs, following standard practices for transformer-based sequence labeling.

\subsection{Evaluation Metrics}

\paragraph{Model Evaluation}
We evaluate model performance using Pairwise F1 \cite{hripcsak_agreement_2005, deleger_building_2012} and Cohen's $\kappa$. For Pairwise F1, we compute token-level Intersection-over-Union (IoU) between each predicted and gold span of the same category. Predicted and gold spans are matched one-to-one using greedy matching in descending IoU order, and a match is counted as correct when IoU $\geq 0.5$. Unmatched predicted spans are false positives, and unmatched gold spans are false negatives. Micro Pairwise F1 pools true positives, false positives, and false negatives across documents, while macro Pairwise F1 averages document-level F1 scores. Cohen's $\kappa$ is computed over binary token labels, and serves as a stricter measure of exact boundary agreement. More details on how this is implemented can be found in Appendix~\ref{app:model_eval_metrics}.

\paragraph{Synthetic Data Evaluation} Since models are trained on synthetic spans, we analyze how well they approximate the gold-standard real spans across two dimensions: structural and lexical. Structural properties include span length and span position within the document. Lexical properties include measures of vocabulary diversity and concentration, such as Moving Average Type-Token Ratio (MATTR) \cite{Covington01052010}, Expectation-Adjusted Distinct-2 (EAD-2) \cite{liu-etal-2022-rethinking}, Top-20 token coverage, and vocabulary per span. The exact definitions and formulation of these metrics can be found in Appendix~\ref{app:synth_metrics}. For each metric, we compute the relative shift between synthetic and real spans, where values closer to zero indicate stronger alignment; positive values indicate over-representation in synthetic spans, while negative values indicate under-representation (see Figure \ref{fig:heatmap_relative_shift}). Metric formulations and further descriptions are also provided in Appendix~\ref{app:synth_metrics}.

\section{Results and Discussion}
\label{sec:results}
\label{sec:analysis}

\subsection{Synthetic Data Quality}
\label{sec:synth_results}

Since encoder models are trained on synthetic spans, we compare them against gold annotations. Synthetic spans mostly match the structural properties of real spans, but \textit{Coherence} shows a structural mismatch, with synthetic spans tending to be longer despite span-length constraints (see Figure \ref{fig:synth_vs_real_structural}). Lexically, synthetic spans are less diverse: EAD-2 decreases across all categories, vocabulary per span drops substantially, and Top-20 token coverage increases most for \textit{Unsupported Claim} and \textit{Lacks Synthesis}. Overall, synthetic data is more reliable for surface-level categories such as \textit{Format} than for discourse-level categories requiring rhetorical variation. Additional analysis is provided in Appendix~\ref{app:synth_correlation}.

\subsection{Model Performance and Task Difficulty}
\label{sec:model_results}

\begin{table*}[t]
\centering
\scriptsize
\setlength{\tabcolsep}{3pt}
\renewcommand{\arraystretch}{1.1}

\resizebox{\textwidth}{!}{
\begin{tabular}{lccc|cccccc|ccc|ccc}
\toprule
& \multicolumn{3}{c|}{}
& \multicolumn{6}{c|}{\textbf{Unsupported Claim}}
& \multicolumn{3}{c|}{}
& \multicolumn{3}{c}{} \\

\cmidrule(lr){5-10}
\textbf{Model}
& \multicolumn{3}{c|}{\textbf{Format}}
& \multicolumn{3}{c}{Short}
& \multicolumn{3}{c|}{Long}
& \multicolumn{3}{c|}{\textbf{Coherence}}
& \multicolumn{3}{c}{\textbf{Lacks Synthesis}} \\

\cmidrule(lr){2-4}
\cmidrule(lr){5-7}
\cmidrule(lr){8-10}
\cmidrule(lr){11-13}
\cmidrule(lr){14-16}

& Micro & Macro & $\kappa$
& Micro & Macro & $\kappa$
& Micro & Macro & $\kappa$
& Micro & Macro & $\kappa$
& Micro & Macro & $\kappa$ \\

\midrule

\textbf{GPT-5 Zero}
& 0.416 & 0.442 & 0.265
& \textbf{0.453} & \underline{0.275} & \underline{0.340}
& 0.441 & 0.310 & 0.120
& 0.286 & 0.322 & \underline{0.094}
& 0.039 & 0.048 & \textbf{0.090} \\

\textbf{GPT-5 Few}
& 0.392 & 0.440 & 0.157
& \underline{0.400} & 0.246 & \textbf{0.351}
& 0.495 & 0.329 & \textbf{0.172}
& 0.278 & 0.294 & \textbf{0.129}
& 0.043 & 0.056 & \underline{0.041} \\

\hline

\textbf{XLM-R}
& \underline{0.648} & \underline{0.609} & \underline{0.598}
& 0.263 & 0.219 & 0.234
& \textbf{0.567} & \textbf{0.462} & 0.099
& 0.810 & 0.769 & 0.014
& \textbf{0.990} & \textbf{0.990} & 0.000 \\

\textbf{SciBERT}
& \textbf{0.667} & \textbf{0.628} & \textbf{0.653}
& 0.335 & \textbf{0.318} & 0.280
& 0.169 & 0.113 & 0.038
& 0.975 & 0.974 & -0.018
& \underline{0.984} & \underline{0.986} & -0.014 \\

\textbf{ModernBERT}
& 0.533 & 0.598 & 0.551
& 0.151 & 0.160 & 0.129
& \underline{0.497} & \underline{0.414} & \underline{0.141}
& \underline{0.985} & \underline{0.981} & -0.015
& 0.708 & 0.755 & 0.005 \\

\textbf{NeoBERT}
& 0.053 & 0.041 & -0.030
& 0.205 & 0.158 & 0.082
& 0.053 & 0.039 & -0.093
& \textbf{0.988} & \textbf{0.987} & -0.009
& 0.816 & 0.762 & 0.022 \\

\bottomrule
\end{tabular}
}

\caption{Model performance across error categories. Micro and macro refer
to Pairwise F1, and $\kappa$ refers to pooled Cohen's $\kappa$.
{\em Unsupported Claim} is split into short and long spans.
{\em Short Unsupported Claim} uses citation-based post-processing.
Best scores are bolded and second-best scores are underlined.}
\label{tab:model_results_combined}
\end{table*}

The results reveal a gap between approximate issue localization and precise boundary agreement. As shown in Table~\ref{tab:model_results_combined}, encoder models often obtain high Pairwise F1, especially for longer or paragraph-level categories, but these gains are not always reflected in Cohen's $\kappa$. This discrepancy arises because Pairwise F1 rewards sufficient span overlap, whereas $\kappa$ measures stricter token-level agreement and is more sensitive to boundary mismatches, over-extended predictions, and class imbalance.

This behavior is most visible for \textit{Coherence} and \textit{Lacks Synthesis}: models often identify the broad region where a discourse-level issue occurs, but do not align with exact gold boundaries. However, as noted earlier, these categories are inherently approximate feedback regions rather than precise token-level errors, so partial boundary overlap is expected and still conveys meaningful diagnostic signal. In contrast, surface-level categories such as \textit{Format} show more stable agreement because they rely on explicit citation-formatting cues.

Prompting-based baselines show a complementary limitation. They can identify broad regions that contain an issue, but often produce spans that are too long or inconsistently bounded, consistent with prior work showing that LLMs struggle with precise span extraction \cite{semin2026strategiesspanlabelinglarge, jarolim2025llmsextracthumanlikefinegrained}. 

Overall, Pairwise F1 and $\kappa$ capture complementary aspects of the task: approximate localization and strict boundary agreement. The usefulness of approximate span feedback remains to be validated through an extrinsic user study in the future.

\section{Conclusion}
\label{sec:conclusion}

We introduced a span-level benchmark for diagnostic feedback on Related Work writing, grounded in a taxonomy of citation- and discourse-based issue categories. The benchmark contributes a human-annotated evaluation set, a synthetic training set, and category-specific evaluation settings for studying writing issues that are local to the text but often depend on citation use, discourse coherence, and authorial positioning.

Our experiments show that existing models capture different aspects of the task. Encoder models often achieve stronger overlap-based localization than prompting baselines, especially for longer or paragraph-level categories, but their low Cohen's $\kappa$ indicates weak exact boundary agreement. Prompting-based baselines are sometimes competitive under $\kappa$ and perform strongly for \textit{Short Unsupported Claim}, but remain unreliable for localized feedback in discourse-level categories such as \textit{Lacks Synthesis}. Surface-level \textit{Format} errors are detected most consistently, while \textit{Short Unsupported Claim} depends strongly on citation-aware constraints rather than model architecture alone.

These findings show why this benchmark is useful: to our knowledge, it is among the first resources to group writing and argumentative issues that are specific to how authors cite, relate, and synthesize prior work in Related Work sections. By separating surface-level citation problems from argumentative and discourse-level writing issues, the benchmark allows future work to evaluate whether models can move beyond detecting local textual patterns toward providing meaningful writing feedback. More broadly, the dataset supplies a testbed for studying the gap between span localization and argument-level reasoning in academic writing. Future work should build on this resource by incorporating explanation generation, stronger citation and discourse modeling, and user studies that evaluate whether span-level diagnostic feedback helps writers revise Related Work sections in practice.

\section*{Limitations}
\label{sec:future_lim}
Synthetic data remains a key limitation, as it lacks the lexical and rhetorical diversity of real writing. Future work should explore hybrid annotation strategies combining model-generated spans with human validation.

Furthermore, we treat GPT-5 zero/few-shot prompting vs. supervised training as a starting point for performance comparison and leave broader and more sophisticated methods that use open-weight models or agentic pipelines for future work. 

Discourse-level categories require deeper reasoning over argument structure. Incorporating argumentation and discourse modeling may improve performance. Additionally, \textit{Short Unsupported Claim} detection currently relies on heuristic post-processing; future work could explore approaches that jointly model citation context.

The framework also lacks interpretability and user-centered evaluation. Extending the system to generate explanations and conducting user studies are important next steps. Finally, multi-task learning may better capture dependencies between error categories.

\section*{Ethics Statement}
Although this work is intended to be an experimental study of modeling approaches for span-level academic writing error detection, rather than a ready-to-deploy writing feedback system, similar approaches could be misused if they are utilized in grading, assessment, or publication without human input or review. We emphasize that users of such systems bear the responsibility of making their own judgment as to whether outputs of such systems are applicable to their work.  

Furthermore, during the annotation process, we used the {\tt NLPeer} dataset which is distributed under CC-BY-NC-SA 4.0 license, for the purpose of studying span-level academic writing error detection. The annotations made in this work are intended to support experimental modeling of this task and to evaluate writing quality in paper submissions. No demographic information about authors was used during the annotation, training or evaluation.

Our annotators participated in this work on a voluntary basis and were informed that they could opt out or discontinue their participation at any point during the annotation process. They were also informed about the purpose of the study, that their annotations would be used to evaluate writing issues in literature reviews or adjacent sections. Annotators were provided with annotation guidelines, and a pilot round was carried out to reconcile differences in opinion before the final annotation round.

Large language models were used in this study to help edit text and debug code. More specifically, they were used to improve readability and grammar by suggesting less repetitive phrasing. Furthermore, they assisted in debugging code and suggesting cleaner implementation of functions. All final decisions were made by the authors, who have full responsibility over the research ideas, task formulation, decisions in methodological design, experiments, and analysis done in this study.

\bibliography{custom}

@misc{cmu_lit_review,
  author = {{Carnegie Mellon University}},
  title = {{Literature Reviews}},
  year = {2026},
  howpublished = {\url{https://www.cmu.edu/student-success/other-resources/resource-descriptions/related-work.html}},
  note = {Accessed: 2026-04-20}
}

@inproceedings{teufel-etal-2006-automatic,
    title = "Automatic classification of citation function",
    author = "Teufel, Simone  and
      Siddharthan, Advaith  and
      Tidhar, Dan",
    editor = "Jurafsky, Dan  and
      Gaussier, Eric",
    booktitle = "Proceedings of the 2006 Conference on Empirical Methods in Natural Language Processing",
    series = {EMNLP~`2006},
    NOmonth = jul,
    year = "2006",
    address = "Sydney, Australia",
    publisher = "Association for Computational Linguistics",
    url = "https://aclanthology.org/W06-1613/",
    pages = "103--110"
}

@inproceedings{li-ouyang-2024-related,
    title = "{Related Work and Citation Text Generation: A Survey}",
    author = "Li, Xiangci  and
      Ouyang, Jessica",
    editor = "Al-Onaizan, Yaser  and
      Bansal, Mohit  and
      Chen, Yun-Nung",
    booktitle = "Proceedings of the 2024 Conference on Empirical Methods in Natural Language Processing",
    month = nov,
    year = "2024",
    address = "Miami, Florida, USA",
    publisher = "Association for Computational Linguistics",
    url = "https://aclanthology.org/2024.emnlp-main.767/",
    doi = "10.18653/v1/2024.emnlp-main.767",
    pages = "13846--13864"
}

@inproceedings{li-etal-2016-diversity,
    title = "{A Diversity-Promoting Objective Function for Neural Conversation Models}",
    author = "Li, Jiwei  and
      Galley, Michel  and
      Brockett, Chris  and
      Gao, Jianfeng  and
      Dolan, Bill",
    editor = "Knight, Kevin  and
      Nenkova, Ani  and
      Rambow, Owen",
    booktitle = "Proceedings of the 2016 Conference of the North {A}merican Chapter of the Association for Computational Linguistics: Human Language Technologies",
    month = jun,
    year = "2016",
    address = "San Diego, California",
    publisher = "Association for Computational Linguistics",
    url = "https://aclanthology.org/N16-1014/",
    doi = "10.18653/v1/N16-1014",
    pages = "110--119"
}

@inproceedings{li-ouyang-2025-explaining,
    title = "{Explaining Relationships Among Research Papers}",
    author = "Li, Xiangci  and
      Ouyang, Jessica",
    editor = "Rambow, Owen  and
      Wanner, Leo  and
      Apidianaki, Marianna  and
      Al-Khalifa, Hend  and
      Eugenio, Barbara Di  and
      Schockaert, Steven",
    booktitle = "Proceedings of the 31st International Conference on Computational Linguistics",
    month = jan,
    year = "2025",
    address = "Abu Dhabi, UAE",
    publisher = "Association for Computational Linguistics",
    url = "https://aclanthology.org/2025.coling-main.73/",
    pages = "1080--1105"
}

@inproceedings{siddharthan-teufel-2007-whose,
    title = "{Whose Idea Was This, and Why Does it Matter? Attributing Scientific Work to Citations}",
    author = "Siddharthan, Advaith  and
      Teufel, Simone",
    editor = "Sidner, Candace  and
      Schultz, Tanja  and
      Stone, Matthew  and
      Zhai, ChengXiang",
    booktitle = "Human Language Technologies 2007: The Conference of the North {A}merican Chapter of the Association for Computational Linguistics; Proceedings of the Main Conference",
    month = apr,
    year = "2007",
    address = "Rochester, New York",
    publisher = "Association for Computational Linguistics",
    url = "https://aclanthology.org/N07-1040/",
    pages = "316--323"
}

@inproceedings{lauscher-etal-2022-multicite,
    title = "{{M}ulti{C}ite: Modeling realistic citations requires moving beyond the single-sentence single-label setting}",
    author = "Lauscher, Anne  and
      Ko, Brandon  and
      Kuehl, Bailey  and
      Johnson, Sophie  and
      Cohan, Arman  and
      Jurgens, David  and
      Lo, Kyle",
    editor = "Carpuat, Marine  and
      de Marneffe, Marie-Catherine  and
      Meza Ruiz, Ivan Vladimir",
    booktitle = "Proceedings of the 2022 Conference of the North American Chapter of the Association for Computational Linguistics: Human Language Technologies",
    month = jul,
    year = "2022",
    address = "Seattle, United States",
    publisher = "Association for Computational Linguistics",
    url = "https://aclanthology.org/2022.naacl-main.137/",
    doi = "10.18653/v1/2022.naacl-main.137",
    pages = "1875--1889"
}

@article{jurgens-etal-2018-measuring,
    title = "{Measuring the Evolution of a Scientific Field through Citation Frames}",
    author = "Jurgens, David  and
      Kumar, Srijan  and
      Hoover, Raine  and
      McFarland, Dan  and
      Jurafsky, Dan",
    editor = "Lee, Lillian  and
      Johnson, Mark  and
      Toutanova, Kristina  and
      Roark, Brian",
    journal = "Transactions of the Association for Computational Linguistics",
    volume = "6",
    year = "2018",
    address = "Cambridge, MA",
    publisher = "MIT Press",
    url = "https://aclanthology.org/Q18-1028/",
    doi = "10.1162/tacl_a_00028",
    pages = "391--406"
}

@article{aburaed_automatic_2020,
	title = {Automatic related work section generation: experiments in scientific document abstracting},
	volume = {125},
	issn = {1588-2861},
	url = {https://doi.org/10.1007/s11192-020-03630-2},
	doi = {10.1007/s11192-020-03630-2},
	number = {3},
	journal = {Scientometrics},
	author = {AbuRa’ed, Ahmed and Saggion, Horacio and Shvets, Alexander and Bravo, Àlex},
	month = dec,
	year = {2020},
	pages = {3159--3185},
}

@article{sahinuc2026rewardmodelingscientificwriting,
  title={{Reward Modeling for Scientific Writing Evaluation}},
  author={{\c{S}}ahinu{\c{c}}, Furkan and Dutta, Subhabrata and Gurevych, Iryna},
  journal={arXiv preprint arXiv:2601.11374},
  year={2026}
}

@article{sahinuc2026expertpreferencebasedevaluationautomated,
      title={{Expert Preference-based Evaluation of Automated Related Work Generation}}, 
      author={{\c{S}}ahinu{\c{c}}, Furkan and Dutta, Subhabrata and Gurevych, Iryna},
      journal={arXiv preprint arXiv:2508.07955},
      year={2025}
}

@misc{openai2025o3mini,
  author = {{OpenAI}},
  title = {{OpenAI o3-mini}},
  year = {2025},
  month = jan,
  url = {https://openai.com/index/openai-o3-mini/},
  note = {Published January 31, 2025. Accessed: 2026-04-28}
}

@inproceedings{zyska2026exposiateachingassessmentacademic,
    title = "{Expos{\'i}a: Teaching and Assessment of Academic Writing Skills for Research Project Proposals and Peer Feedback}",
    author = "Zyska, Dennis  and
      Rozovskaya, Alla  and
      Kuznetsov, Ilia  and
      Gurevych, Iryna",
    editor = "Lee, Hung-Yi  and
      Dev, Sunipa  and
      Li, Junyi Jessy  and
      Elliott, Desmond",
    booktitle = "Proceedings of the 2026 Conference on Empirical Methods in Natural Language Processing",
    month = oct,
    year = "2026",
    address = "Budapest, Hungary",
    publisher = "Association for Computational Linguistics",
    url = "https://aclanthology.org/2026.emnlp-main.49/",
    doi = "10.18653/2026.emnlp-main.49",
    pages = "1076--1130",
    ISBN = "979-8-89176-459-0"
}

@inproceedings{dycke-etal-2023-nlpeer,
    title = "{{NLP}eer: A Unified Resource for the Computational Study of Peer Review}",
    author = "Dycke, Nils  and
      Kuznetsov, Ilia  and
      Gurevych, Iryna",
    editor = "Rogers, Anna  and
      Boyd-Graber, Jordan  and
      Okazaki, Naoaki",
    booktitle = "Proceedings of the 61st Annual Meeting of the Association for Computational Linguistics (Volume 1: Long Papers)",
    month = jul,
    year = "2023",
    address = "Toronto, Canada",
    publisher = "Association for Computational Linguistics",
    url = "https://aclanthology.org/2023.acl-long.277/",
    doi = "10.18653/v1/2023.acl-long.277",
    pages = "5049--5073"
}

@inproceedings{Hoang2010TowardsAR,
    title={{Towards Automated Related Work Summarization}},
    author={Hoang, Cong Duy Vu and Kan, Min-Yen},
    booktitle={Coling 2010: Posters},
    pages={427--435},
    year={2010}
}

@inproceedings{kan2001informative,
  title     = {{Domain-specific Informative and Indicative Summarization for Information Retrieval}},
  author    = {Kan, Min-Yen and McKeown, Kathleen R. and Klavans, Judith L.},
  booktitle = {Proceedings of the 2001 ACM CIKM Workshop on Information Retrieval},
  year      = {2001}
}

@article{Yasunaga_Kasai_Zhang_Fabbri_Li_Friedman_Radev_2019, 
    title={{ScisummNet: A Large Annotated Corpus and Content-Impact Models for Scientific Paper Summarization with Citation Networks}}, 
    volume={33}, 
    url={https://ojs.aaai.org/index.php/AAAI/article/view/4727}, 
    DOI={10.1609/aaai.v33i01.33017386}, 
    abstractNote={&lt;p&gt;Scientific article summarization is challenging: large, annotated corpora are not available, and the summary should ideally include the article’s impacts on research community. This paper provides novel solutions to these two challenges. We 1) develop and release the first large-scale manually-annotated corpus for scientific papers (on computational linguistics) by enabling faster annotation, and 2) propose summarization methods that integrate the authors’ original highlights (abstract) and the article’s actual impacts on the community (citations), to create comprehensive, hybrid summaries. We conduct experiments to demonstrate the efficacy of our corpus in training data-driven models for scientific paper summarization and the advantage of our hybrid summaries over abstracts and traditional citation-based summaries. Our large annotated corpus and hybrid methods provide a new framework for scientific paper summarization research.&lt;/p&gt;}, 
    number={01}, 
    journal={Proceedings of the AAAI Conference on Artificial Intelligence}, 
    author={Yasunaga, Michihiro and Kasai, Jungo and Zhang, Rui and Fabbri, Alexander R. and Li, Irene and Friedman, Dan and Radev, Dragomir R.}, 
    year={2019}, 
    month={Jul.}, 
    pages={7386-7393} }

@inproceedings{chen-etal-2021-capturing,
    title = "{Capturing Relations between Scientific Papers: An Abstractive Model for Related Work Section Generation}",
    author = "Chen, Xiuying  and
      Alamro, Hind  and
      Li, Mingzhe  and
      Gao, Shen  and
      Zhang, Xiangliang  and
      Zhao, Dongyan  and
      Yan, Rui",
    editor = "Zong, Chengqing  and
      Xia, Fei  and
      Li, Wenjie  and
      Navigli, Roberto",
    booktitle = "Proceedings of the 59th Annual Meeting of the Association for Computational Linguistics and the 11th International Joint Conference on Natural Language Processing (Volume 1: Long Papers)",
    month = aug,
    year = "2021",
    address = "Online",
    publisher = "Association for Computational Linguistics",
    url = "https://aclanthology.org/2021.acl-long.473/",
    doi = "10.18653/v1/2021.acl-long.473",
    pages = "6068--6077"
}

@article{chen-automatic-gen-related-works,
  title={Automatic generation of related work through summarizing citations},
  author={Chen, Jingqiang and Zhuge, Hai},
  journal={Concurrency and Computation: Practice and Experience},
  volume={31},
  number={3},
  pages={e4261},
  year={2019},
  publisher={Wiley Online Library},
  doi = {10.1002/cpe.4261}
}

@online{OpenAI2025GPT5,
  author       = {{OpenAI}},
  title        = {{GPT-5}},
  year         = {2025},
  url          = {https://openai.com/gpt-5/},
  note         = {Accessed: 2026-01-27},
  organization = {OpenAI}
}

@inproceedings{xlm-roberta,
    title = "{Unsupervised Cross-lingual Representation Learning at Scale}",
    author = "Conneau, Alexis  and
      Khandelwal, Kartikay  and
      Goyal, Naman  and
      Chaudhary, Vishrav  and
      Wenzek, Guillaume  and
      Guzm{\'a}n, Francisco  and
      Grave, Edouard  and
      Ott, Myle  and
      Zettlemoyer, Luke  and
      Stoyanov, Veselin",
    editor = "Jurafsky, Dan  and
      Chai, Joyce  and
      Schluter, Natalie  and
      Tetreault, Joel",
    booktitle = "Proceedings of the 58th Annual Meeting of the Association for Computational Linguistics",
    month = jul,
    year = "2020",
    address = "Online",
    publisher = "Association for Computational Linguistics",
    url = "https://aclanthology.org/2020.acl-main.747/",
    doi = "10.18653/v1/2020.acl-main.747",
    pages = "8440--8451"
}

@inproceedings{scibert,
    title = "{{S}ci{BERT}: A Pretrained Language Model for Scientific Text}",
    author = "Beltagy, Iz  and
      Lo, Kyle  and
      Cohan, Arman",
    editor = "Inui, Kentaro  and
      Jiang, Jing  and
      Ng, Vincent  and
      Wan, Xiaojun",
    booktitle = "Proceedings of the 2019 Conference on Empirical Methods in Natural Language Processing and the 9th International Joint Conference on Natural Language Processing (EMNLP-IJCNLP)",
    month = nov,
    year = "2019",
    address = "Hong Kong, China",
    publisher = "Association for Computational Linguistics",
    url = "https://aclanthology.org/D19-1371/",
    doi = "10.18653/v1/D19-1371",
    pages = "3615--3620"
}

@inproceedings{modernbert,
    title = "{Smarter, Better, Faster, Longer: A Modern Bidirectional Encoder for Fast, Memory Efficient, and Long Context Finetuning and Inference}",
    author = {Warner, Benjamin  and
      Chaffin, Antoine  and
      Clavi{\'e}, Benjamin  and
      Weller, Orion  and
      Hallstr{\"o}m, Oskar  and
      Taghadouini, Said  and
      Gallagher, Alexis  and
      Biswas, Raja  and
      Ladhak, Faisal  and
      Aarsen, Tom  and
      Adams, Griffin Thomas  and
      Howard, Jeremy  and
      Poli, Iacopo},
    editor = "Che, Wanxiang  and
      Nabende, Joyce  and
      Shutova, Ekaterina  and
      Pilehvar, Mohammad Taher",
    booktitle = "Proceedings of the 63rd Annual Meeting of the Association for Computational Linguistics (Volume 1: Long Papers)",
    month = jul,
    year = "2025",
    address = "Vienna, Austria",
    publisher = "Association for Computational Linguistics",
    url = "https://aclanthology.org/2025.acl-long.127/",
    doi = "10.18653/v1/2025.acl-long.127",
    pages = "2526--2547",
    ISBN = "979-8-89176-251-0"
}

@article{neobert,
      title={{NeoBERT: A Next-Generation BERT}}, 
        author={Breton, Lola Le and Fournier, Quentin and Mezouar, Mariam El and Morris, John X and Chandar, Sarath},
        journal={arXiv preprint arXiv:2502.19587},
        year={2025},
        url={https://arxiv.org/abs/2502.19587},
}

@article{jarolim2025llmsextracthumanlikefinegrained,
  title={{Can LLMs extract human-like fine-grained evidence for evidence-based fact-checking?}},
  author={Jarol{\'\i}m, Anton{\'\i}n and Faj{\v{c}}{\'\i}k, Martin and Makaiov{\'a}, Lucia},
  journal={arXiv preprint arXiv:2511.21401},
  year={2025}
}

@article{semin2026strategiesspanlabelinglarge,
  title={{Strategies for Span Labeling with Large Language Models}},
  author={Semin, Danil and Du{\v{s}}ek, Ond{\v{r}}ej and Kasner, Zden{\v{e}}k},
  journal={arXiv preprint arXiv:2601.16946},
  year={2026}
}

@article{hripcsak_agreement_2005,
	address = {England},
	title = {{Agreement, the F-measure, and reliability in information retrieval}},
	volume = {12},
	issn = {1067-5027 1527-974X},
	doi = {10.1197/jamia.M1733},
	language = {eng},
	number = {3},
	journal = {Journal of the American Medical Informatics Association : JAMIA},
	author = {Hripcsak, George and Rothschild, Adam S.},
	month = jun,
	year = {2005},
	pages = {296--298},
}

@article{deleger_building_2012,
	address = {United States},
	title = {Building gold standard corpora for medical natural language processing tasks},
	volume = {2012},
	issn = {1942-597X 1559-4076},
	language = {eng},
	journal = {AMIA Annual Symposium proceedings. AMIA Symposium},
	author = {Deleger, Louise and Li, Qi and Lingren, Todd and Kaiser, Megan and Molnar, Katalin and Stoutenborough, Laura and Kouril, Michal and Marsolo, Keith and Solti, Imre},
	year = {2012},
	pages = {144--153},
}

@article{Covington01052010,
    author = {Michael A. Covington and Joe D. McFall},
    title = {{Cutting the Gordian Knot: The Moving-Average Type–Token Ratio (MATTR)}},
    journal = {Journal of Quantitative Linguistics},
    volume = {17},
    number = {2},
    pages = {94--100},
    year = {2010},
    publisher = {Routledge},
    doi = {10.1080/09296171003643098},
    URL = {https://doi.org/10.1080/09296171003643098},
    eprint = {https://doi.org/10.1080/09296171003643098}
}

@inproceedings{liu-etal-2022-rethinking,
    title = "{Rethinking and Refining the Distinct Metric}",
    author = "Liu, Siyang  and
      Sabour, Sahand  and
      Zheng, Yinhe  and
      Ke, Pei  and
      Zhu, Xiaoyan  and
      Huang, Minlie",
    editor = "Muresan, Smaranda  and
      Nakov, Preslav  and
      Villavicencio, Aline",
    booktitle = "Proceedings of the 60th Annual Meeting of the Association for Computational Linguistics (Volume 2: Short Papers)",
    month = may,
    year = "2022",
    address = "Dublin, Ireland",
    publisher = "Association for Computational Linguistics",
    url = "https://aclanthology.org/2022.acl-short.86/",
    doi = "10.18653/v1/2022.acl-short.86",
    pages = "762--770"
}

@inproceedings{luan-etal-2018-multi,
    title = "{Multi-Task Identification of Entities, Relations, and Coreference for Scientific Knowledge Graph Construction}",
    author = "Luan, Yi  and
      He, Luheng  and
      Ostendorf, Mari  and
      Hajishirzi, Hannaneh",
    editor = "Riloff, Ellen  and
      Chiang, David  and
      Hockenmaier, Julia  and
      Tsujii, Jun{'}ichi",
    booktitle = "Proceedings of the 2018 Conference on Empirical Methods in Natural Language Processing",
    month = oct # "-" # nov,
    year = "2018",
    address = "Brussels, Belgium",
    publisher = "Association for Computational Linguistics",
    url = "https://aclanthology.org/D18-1360/",
    doi = "10.18653/v1/D18-1360",
    pages = "3219--3232"
}

@inproceedings{kocmi-federmann-2023-gemba-mqm,
    title = {{GEMBA-MQM: Detecting Translation Quality Error Spans with GPT-4}},
    author = {Kocmi, Tom  and Federmann, Christian},
    booktitle = "Proceedings of the Eighth Conference on Machine Translation",
    month = dec,
    year = "2023",
    address = "Singapore",
    publisher = "Association for Computational Linguistics",
}

@article{guerreiro-etal-2024-xcomet,
    title = "{x{COMET}: Transparent Machine Translation Evaluation through Fine-grained Error Detection}",
    author = "Guerreiro, Nuno M.  and
      Rei, Ricardo  and
      van Stigt, Daan  and
      Coheur, Luisa  and
      Colombo, Pierre  and
      Martins, Andr{\'e} F. T.",
    journal = "Transactions of the Association for Computational Linguistics",
    volume = "12",
    year = "2024",
    address = "Cambridge, MA",
    publisher = "MIT Press",
    url = "https://aclanthology.org/2024.tacl-1.54/",
    doi = "10.1162/tacl_a_00683",
    pages = "979--995"
}

\appendix

\begin{table*}[t]
\centering
\footnotesize
\setlength{\tabcolsep}{5pt}
\renewcommand{\arraystretch}{1.15}

\begin{tabularx}{\textwidth}{p{0.16\textwidth} X X}
\toprule
\textbf{Category} & \textbf{Example} & \textbf{Explanation} \\
\midrule

\textit{Format}
&
``Following the taxonomy \uline{in (Liu et al., 2021)}, these methods can be classified depending on whether the language models' parameters are tuned and on whether trainable prompts are introduced.''
&
The highlighted span contains an incorrect citation format. The citation is written in parenthetical form even though the sentence requires a narrative citation. The correct form would be ``Following the taxonomy in Liu et al.\ (2021)''. \\

\midrule

\textit{Unsupported Claim}
&
``Following the work of DrQA (Chen et al., 2017), \uline{most recent works build a two-stage retriever-reader system to tackle the problem.} The retriever aims at retrieving supportive passages to the given question from a large document corpus. The reader intends to find answer of the question from the first stage retrieved passages.''
&
The highlighted claim refers to ``most recent works'' but does not provide citations to the works being discussed. Since the statement makes a claim about prior research trends, it requires explicit attribution to supporting studies. \\

\midrule

\textit{Coherence}
&
\uline{``For instance, Mao et al.\ (2019) generates story with multitasking learning on commonsense QA datasets. Zhao et al.\ (2020) used BERT as a knowledge selection module for dialogue generation. Chakrabarty et al.\ (2020) ranked knowledge generated from the COMET for sarcasm generation. Ji et al.\ (2020) do multi-hop with a graph convolutional network on ConceptNet.''}
&
The highlighted span lists several cited works sequentially, but does not explain how they are related to each other or why they are grouped together. The paragraph informs the reader that these papers exist, but does not provide a transition, comparison, or clear narrative relation between them. \\

\midrule

\textit{Lacks Synthesis}
&
\uline{``Other proposed value schemes are more generic. Combining research from anthropology, sociology, philosophy, and psychology, Rokeach (1973) estimates the total number of human values to be fewer than hundreds, and develops a practical survey of 36 values that distinguishes between values pertaining to desirable end states and desirable behavior. Specifically for cross-cultural analyses, Schwartz et al.\ (2012) derived 48 value questions from the universal needs of individuals and societies, including obeying all the laws and to be humble. Moreover, Schwartz (1994) proposes a relatedness of values by their tendency to be compatible in their pursuit.''}
&
The highlighted paragraph summarizes several prior works but does not connect them to the author's own argument, contribution, or research gap. It is unclear why these studies are discussed together or how they motivate the current work. As a result, the paragraph remains descriptive rather than synthetic or argumentative. \\

\bottomrule
\end{tabularx}

\caption{Representative examples from the annotation guidelines. Underlined text indicates the annotated span.}
\label{tab:category_examples}
\end{table*}

\section{Annotation Guidelines}
\label{app:annotation_guidelines}

Annotators were asked to identify span-level writing issues in short excerpts from the Introduction and Related Work sections of academic papers. The task focused on four error categories: \textit{Format}, \textit{Unsupported Claim}, \textit{Coherence}, and \textit{Lacks Synthesis}. Annotators were instructed to highlight only spans that matched the category definitions and to leave passages unannotated when no error was present.

Because the categories operate at different linguistic levels, the guidelines specified category-specific span boundaries. \textit{Format} errors were annotated at the citation level, \textit{Unsupported Claim} errors at the phrase or sentence level, and \textit{Coherence} and \textit{Lacks Synthesis} at the multi-sentence or paragraph level.

\subsection{Category Definitions \& Rules}

\begin{table*}[t]
\centering
\footnotesize
\setlength{\tabcolsep}{5pt}
\renewcommand{\arraystretch}{1.15}

\begin{tabularx}{\textwidth}{
p{0.30\textwidth}
>{\raggedright\arraybackslash}X
>{\raggedright\arraybackslash}X
>{\raggedright\arraybackslash}X
>{\raggedright\arraybackslash}X
}
\toprule
\textbf{In-text Example} & \textbf{A1} & \textbf{A2} & \textbf{A3} & \textbf{A4} \\
\midrule

``Pre-trained multilingual masked language models (PLMs) such as multilingual BERT or XLM-RoBERTa (XLM-R) (Conneau et al., 2020)...''
& \textemdash
& ``multilingual BERT''
& ``multilingual BERT''
& ``multilingual BERT'' \\

\midrule

``In contrast to most prior work, we also show that these results can also be achieved for languages other than English. We compare the hypothesis patterns commonly used in the literature and using the plain label name (null hypothesis) and find that on average there is no significant difference in performance.''
& ``In contrast to most prior work''
& ``In contrast to most prior work, we also show that these results can also be achieved for languages other than English.''
& ``In contrast to most prior work,''
& ``In contrast to most prior work, we also show that these results can also be achieved for languages other than English.'' \\

\bottomrule
\end{tabularx}

\caption{Examples of annotator disagreement for \textit{Unsupported Claim} spans. A1--A4 denote individual annotators.}
\label{tab:ann_examples}
\end{table*}

\paragraph{Format}
Annotators marked citation formatting errors, including missing brackets, missing years, incorrect parenthetical or narrative citation style, and incomplete references. Only the citation itself was annotated, not the surrounding sentence. Format issues usually cover a few words, subsentences, phrases, or segments, rather than whole sentences or multiple sentences.

\paragraph{Unsupported Claim}
Annotators marked claims or mentions of prior work that require citation support. This includes uncited mentions of datasets, models, methods, shared tasks, or vague references such as "recent works". If the unsupported mention was self-contained, for example: "BERT" or "VC Task", the annotators marked the phrase; otherwise, they marked the full sentence. Authors should provide relevant citations at every first mention of a study, paper, shared task, competition, or dataset. Unsupported claims occur as sub-sentences, phrases, segments, or as single sentences. They normally do not occur as multiple sentences. If multiple sentences qualify as being an Unsupported Claim, annotators were asked to highlight the sentences separately.

\paragraph{Coherence}
Annotators marked spans where the relationship between cited works was unclear or where transitions between works were missing. These errors typically occurred across multiple sentences, where papers were listed sequentially without an explicit relation or as a whole paragraph. Annotators were informed to annotate on the multiple sentence or paragraph level. 

\paragraph{Lacks Synthesis}
Annotators marked paragraphs or subsections that described prior work without articulating the author's own perspective, contribution, or research gap. A lack of argument/opinion in the first paragraph is permissible as it serves to be the foundation of the author's argument. Annotations for this category were restricted to multi-sentence spans and did not apply to introductory background paragraphs. 

\subsection{Annotation Workflow}
Annotators were instructed to apply the following checks:

\begin{itemize}
    \setlength{\itemsep}{0pt}
    \item Is the citation formatted correctly? (\textit{Format})
    \item Is each claim or mention of prior work supported by an explicit citation? (\textit{Unsupported Claim})
    \item Are relationships between cited works clearly expressed? (\textit{Coherence})
    \item Does the paragraph explain how prior work relates to the author's own contribution or research gap? (\textit{Lacks Synthesis})
\end{itemize}

Annotators were instructed not to create overlapping annotations at different span levels for the same issue. When uncertain, annotators were asked to annotate only spans they considered sufficiently clear instances of the category.

\subsection{Category Examples}
Examples and explanations for each category can be found in Table~\ref{tab:category_examples}.

\subsection{Annotation Examples}
\label{app:ann_examples}
Table~\ref{tab:ann_examples} provides examples of span annotations made by different annotators for the same input text. These examples illustrate common sources of disagreement in the annotation task, particularly differences in span boundary selection. In some cases, annotators identified the same underlying issue but selected spans of different lengths, while in other cases, an annotator did not mark a span. This reflects the subjective nature of span-level annotation, especially for categories such as \textit{Unsupported Claim}, where determining the exact extent of the unsupported statement can vary across annotators.

\subsection{Annotation Platform}
\begin{figure*}[]
    \centering
    \includegraphics[width=0.9\linewidth]{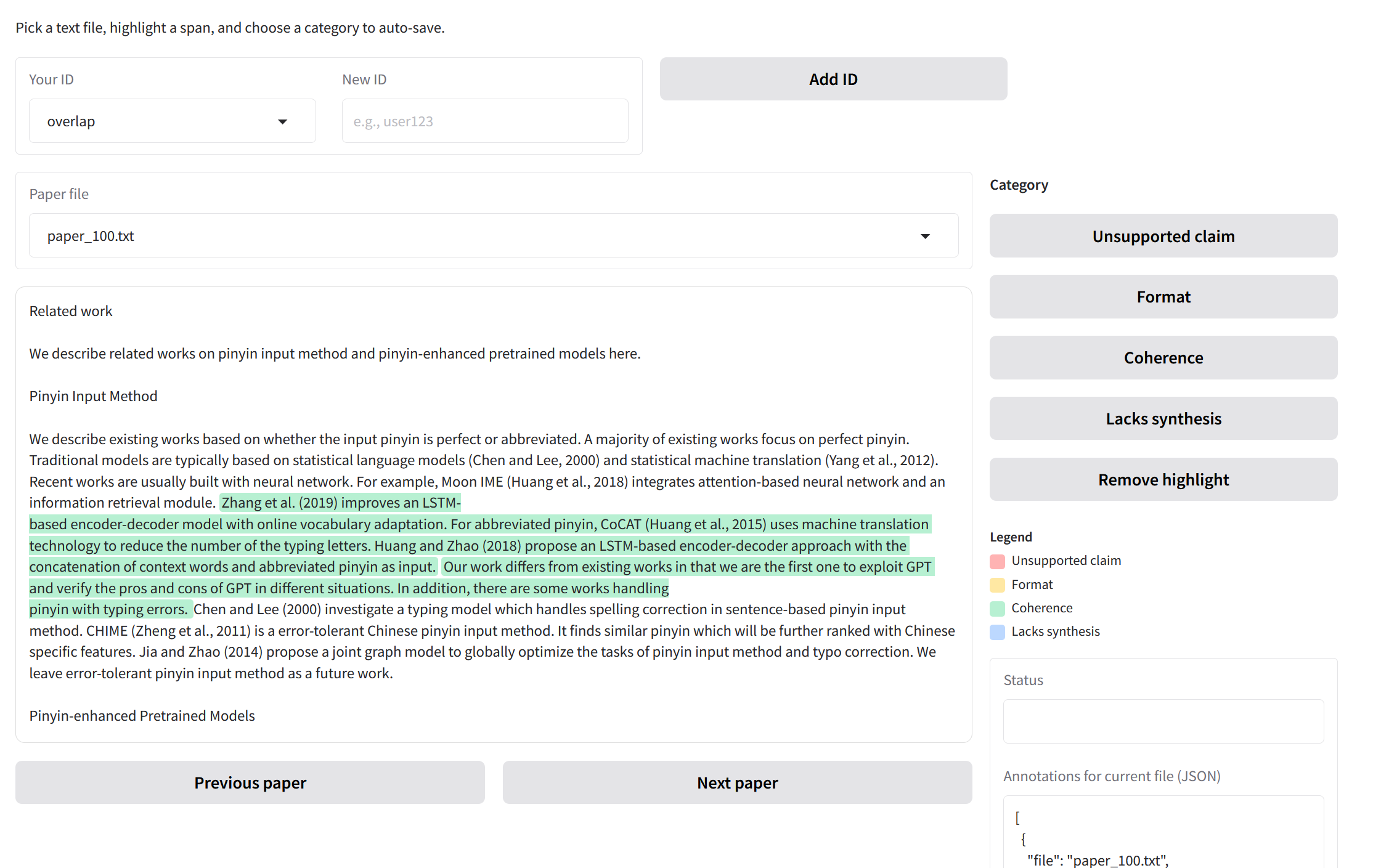}
    \caption{An overview of the annotation interface developed for this study. Annotators are presented with a passage of text to highlight and label spans of text using the categories to the right of the text box.}
    \label{fig:citation_tool}
\end{figure*}

Figure~\ref{fig:citation_tool} shows the annotation interface developed and used in this study. Annotators were presented with the Related Work or Introduction sections of a paper and asked to highlight spans of text that correspond to one of the error categories. The panel on the right displays the categories in bold text to assign the highlighted text to. The start and end characters of the spans, text, label and filename were recorded. 

\section{Synthetic Data Examples}
Table~\ref{tab:synthetic_examples} provides examples of generated synthetic spans for each error category. The examples illustrate how synthetic errors are inserted into real document contexts while preserving the category-specific span granularity. 

\begin{table*}[t]
\centering
\scriptsize
\setlength{\tabcolsep}{5pt}
\renewcommand{\arraystretch}{1.15}

\begin{tabularx}{\textwidth}{
p{0.16\textwidth}
>{\raggedright\arraybackslash}X
>{\raggedright\arraybackslash}X
}
\toprule
\textbf{Category} & \textbf{Synthetic Span} & \textbf{Rationale} \\
\midrule

\textit{Unsupported Claim}
&
``recent studies on DocRE overwhelmingly adopt a two-stage retriever--reader pipeline with sentence-level evidence supervision, reporting gains on DocRED under distant supervision rather than fully annotated settings.''
&
The span asserts a field-wide trend about prior work without providing citations. It therefore represents an unsupported claim about existing research. \\

\midrule

\textit{Format}
&
``Yao et al.''; ``(Devlin et al., 2019; Vaswani et al., 2017''
&
The first span is a narrative citation missing the publication year. The second span is a parenthetical citation with a missing closing parenthesis. In both cases, the error is restricted to the citation span itself. \\

\midrule

\textit{Coherence}
&
``Zhou et al.\ (2021) argue that the Transformer attentions are able to extract useful contextual features across sentences for DocRE, and they adopt an adaptive threshold for each entity pair. Zhang et al.\ (2021) model DocRE as a semantic segmentation task and predict an entity-level relation matrix to capture local and global information. Yao et al.\ (2019) introduce a distant supervision corpus for document-level relations and evaluate sentence aggregation strategies. Christopoulou et al.\ (2019) design an edge-oriented graph that connects mentions and entities with typed links, reporting gains on cross-sentence benchmarks.''
&
The span lists several cited works sequentially without explaining how they relate to each other or to the surrounding discussion. This creates an abrupt sequence of citations with unclear transitions between works. \\

\midrule

\textit{Lacks Synthesis}
&
``Prior work has also explored document-level reasoning with heterogeneous supervision sources and evaluation settings, including distant supervision (Mintz et al., 2009; Hoffmann et al., 2011; Zeng et al., 2015), partially annotated or weakly labeled corpora (Zhang et al., 2017; Luo et al., 2020), human-curated benchmarks such as DocRED and CDR (Yao et al., 2019; Li et al., 2016), and large-scale pre-training on encyclopedic text (Petroni et al., 2019; Xiong et al., 2020; Lee et al., 2020; Yamada et al., 2020).''
&
The span enumerates prior works, datasets, and methods without connecting them to the author's argument, contribution, or research gap. It remains descriptive rather than synthetic or argumentative. \\

\bottomrule
\end{tabularx}

\caption{Examples of synthetic spans generated for each error category. The table shows the generated span and the rationale used to validate its category assignment.}
\label{tab:synthetic_examples}
\end{table*}

\section{Synthetic Data Evaluation Metrics}
\label{app:synth_metrics}

This section describes the metrics used to compare the synthetic spans with the gold-standard human annotations. The goal of this analysis is to assess whether synthetic spans approximate real spans not only structurally, but also lexically. Metrics are computed separately for each error category.

For each metric, we report the relative percentage change between synthetic and real spans:

\begin{align}
    \Delta \% = 
    \frac{\textit{synthetic} - \textit{real}}{\textit{real}} \times 100
\end{align}

\noindent Values close to zero indicate stronger alignment between synthetic and real spans. Negative values indicate that synthetic spans under-represent the property compared to real spans, while positive values indicate over-representation.

\subsection{Structural Metrics}

\paragraph{Span Length}
Span length measures the number of tokens contained in an annotated span. This metric is used to determine whether synthetic spans are similar in size to real annotated spans. Matching span length is important because span-level models learn not only whether an error is present, but also the typical extent of the error in text.

For a span \(s\) with token start index \(\text{start}(s)\) and end index \(\text{end}(s)\), span length is defined as:

\begin{align}
    \mathrm{Length}(s) = \text{end}(s) - \text{start}(s)
\end{align}

\noindent Longer spans indicate broader error regions, while shorter spans indicate more localized errors. This is particularly relevant because the error categories differ in their expected granularity: \textit{Format} errors are usually short citation-level spans, whereas \textit{Coherence} and \textit{Lacks Synthesis} often occur over multiple sentences.

\paragraph{Span Position}
Span position measures where an error span occurs within a document. This is calculated using the normalized center point of a span, allowing comparison across documents of different lengths.

\begin{align}
    \mathrm{Center}(s) =
    \frac{\text{start}(s) + \text{end}(s)}{2L}
\end{align}

\noindent where \(L\) is the total number of tokens in the document. Values closer to \(0\) indicate spans near the beginning of the document, while values closer to \(1\) indicate spans near the end. This metric is included to assess whether synthetic spans occur in similar document regions as real spans.

\subsection{Lexical Metrics}

\paragraph{Moving Average Type-Token Ratio (MATTR)}
MATTR \cite{Covington01052010} measures lexical diversity within a span. It is based on Type-Token Ratio (TTR), which measures the proportion of unique tokens in a text:

\begin{align}
    \mathrm{TTR} =
    \frac{\text{number of unique tokens}}{\text{total number of tokens}}
\end{align}

\noindent However, standard TTR is highly sensitive to text length: shorter texts tend to have artificially high TTR values, while longer texts tend to have lower values because repeated tokens become more likely. MATTR addresses this issue by computing TTR over fixed-size sliding windows and averaging the result.

For a span containing \(N\) tokens and a window size \(w\), MATTR is defined as:

\begin{align}
\mathrm{MATTR}(w)
&= \frac{1}{N-w+1}
\sum_{i=1}^{N-w+1} \mathrm{TTR}_i, \\
\mathrm{TTR}_i
&= \frac{V(i,i+w-1)}{w}.
\end{align}

\noindent where \(V(i, i+w-1)\) denotes the number of unique tokens in the window from position \(i\) to \(i+w-1\). Higher MATTR values indicate greater lexical diversity. In this study, MATTR is used to compare whether synthetic spans contain a similar level of vocabulary variation as real spans.

Because span lengths vary across error categories, the window size \(w\) is set according to the typical span length observed in the gold-standard data: 5 tokens for \textit{Format}, 10 tokens for \textit{Unsupported Claim}, 20 tokens for \textit{Coherence}, and 50 tokens for \textit{Lacks Synthesis}.

\paragraph{Expectation-Adjusted Distinct (EAD)}
Expectation-Adjusted Distinct (EAD) \cite{liu-etal-2022-rethinking} is a length-adjusted version of the Distinct metric \cite{li-etal-2016-diversity}. The original Distinct metric measures the ratio of unique \(n\)-grams to the total number of \(n\)-grams:

\begin{align}
    \mathrm{Distinct} =
    \frac{U}{C}
\end{align}

\noindent where \(U\) is the number of unique \(n\)-grams and \(C\) is the total number of \(n\)-grams. Although useful, this metric is biased by length: as a span becomes longer, the denominator increases, and the probability of repeated \(n\)-grams also increases.

EAD corrects for this by comparing the observed number of unique \(n\)-grams to the expected number of unique \(n\)-grams under a uniform vocabulary distribution. The expected number of unique \(n\)-grams is:

\begin{align}
    \mathbb{E}\left[\hat{U}\right] =
    V\left[
    1 - \left(\frac{V-1}{V}\right)^C
    \right]
\end{align}

\noindent where \(V\) is the vocabulary size and \(C\) is the total number of \(n\)-grams. The EAD score is then:

\begin{align}
    \mathrm{EAD} =
    \frac{U}
    {V\left[
    1 - \left(\frac{V-1}{V}\right)^C
    \right]}
\end{align}

\noindent In this work, EAD-2 is used, meaning the metric is computed over bigrams. This captures phrase-level lexical diversity. Lower EAD-2 values indicate that synthetic spans rely on a smaller or more repetitive set of bigram patterns compared to real spans.

\paragraph{Top-20 Token Coverage}
Top-20 token coverage measures how much of a span is composed of the most frequent tokens. This metric captures lexical concentration: even if a span has reasonable lexical diversity overall, it may still rely heavily on a small set of repeated words or scaffold phrases.

Let \(V_{20}\) be the set of the 20 most frequent tokens in the dataset. For a span with \(N\) tokens, Top-20 token coverage is defined as:

\begin{align}
\mathrm{Top20}(s) =
\frac{\sum_{i=1}^{N} \mathbf{1}(w_i \in V_{20})}{N}
\end{align}

\noindent where \(w_i\) is the \(i\)-th token in the span and \(\mathbf{1}(\cdot)\) is an indicator function. Higher values indicate that a larger proportion of the span consists of highly frequent tokens. This metric is included to identify whether synthetic spans overuse common tokens or repeated phrases.

\begin{table*}[t]
\centering
\small
\setlength{\tabcolsep}{6pt}
\renewcommand{\arraystretch}{1.1}

\begin{tabular}{lccc ccc}
\toprule
\textbf{Model}
& \multicolumn{3}{c}{\textbf{Coherence}}
& \multicolumn{3}{c}{\textbf{Lacks Synthesis}} \\
\cmidrule(lr){2-4}
\cmidrule(lr){5-7}
& $\boldsymbol{t}$ & \textbf{Window} & \textbf{Stride}
& $\boldsymbol{t}$ & \textbf{Window} & \textbf{Stride} \\
\midrule
SciBERT     & 0.3 & 192 & 96 & 0.3 & 192 & 64 \\
ModernBERT  & 0.3 & 192 & 64 & 0.3 & 192 & 64 \\
XLM-R       & 0.3 & 160 & 80 & 0.3 & 192 & 80 \\
NeoBERT     & 0.5 & 160 & 80 & 0.5 & 160 & 80 \\
\bottomrule
\end{tabular}

\caption{Selected hyperparameter configurations for paragraph-level categories. The table reports the coverage threshold $t$, window size, and stride selected for each model.}
\label{tab:hyperparameter_par}
\end{table*}

\begin{table}[t]
\centering
\small
\setlength{\tabcolsep}{6pt}
\renewcommand{\arraystretch}{1.1}

\begin{tabular}{lccc ccc}
\toprule
\textbf{Model}
& \multicolumn{3}{c}{\textbf{Format}}
& \multicolumn{3}{c}{\shortstack{\scriptsize\textbf{Long Unsupported}\\\scriptsize\textbf{Claim}}} \\
\cmidrule(lr){2-4}
\cmidrule(lr){5-7}
& $\boldsymbol{B}$ & $\boldsymbol{I}$ & $\boldsymbol{O}$
& $\boldsymbol{B}$ & $\boldsymbol{I}$ & $\boldsymbol{O}$ \\
\midrule
SciBERT     & 1.0 & 5.0 & 2.0 & 2.0 & 8.0 & 0.5 \\
ModernBERT  & 1.0 & 5.0 & 1.5 & 2.0 & 8.0 & 0.5 \\
XLM-R       & 1.5 & 3.0 & 2.0 & 3.0 & 8.0 & 0.75 \\
NeoBERT     & 1.5 & 3.0 & 1.0 & 3.0 & 6.0 & 0.5 \\
\bottomrule
\end{tabular}

\caption{Selected BIO label weight configurations for token-level categories. The table reports the selected weights for \textit{Format} and long \textit{Unsupported Claim} spans.}
\label{tab:hyperparameter_sent}
\end{table}

\begin{table}[t]
\centering
\small
\setlength{\tabcolsep}{6pt}
\renewcommand{\arraystretch}{1.1}

\begin{tabular}{lcc}
\toprule
\textbf{Model} & \textbf{Binary} & \textbf{Singleton} \\
\midrule
ModernBERT & 0.55 & 0.50 \\
SciBERT    & 0.55 & 0.75 \\
XLM-R      & 0.25 & 0.60 \\
NeoBERT    & 0.55 & 0.65 \\
\bottomrule
\end{tabular}

\caption{Selected threshold values for short \textit{Unsupported Claim} span prediction. The binary threshold determines whether a token is included in a candidate span, while the singleton threshold determines whether isolated predictions are retained.}
\label{tab:hyperparameter_short}
\end{table}

\paragraph{Vocabulary per Span}
Vocabulary per span measures the average number of unique tokens within each span. Unlike MATTR, which normalizes diversity using fixed-size windows, this metric directly measures the amount of lexical variety present in individual spans.

For a set of \(M\) spans, vocabulary per span is defined as:

\begin{align}
    \mathrm{VocabPerSpan} =
    \frac{1}{M}
    \sum_{j=1}^{M}
    |\mathrm{Unique}(s_j)|
\end{align}

\noindent where \(\mathrm{Unique}(s_j)\) is the set of unique tokens in span \(s_j\). Higher values indicate that spans contain a larger number of distinct tokens. This metric is used to assess whether synthetic spans are lexically constrained compared to real spans.

\subsection{Synthetic Training Set Manual Audit}
\label{app:silver_audit}

To assess the quality of the critic-filtered synthetic training data, we manually audited 100 synthetic spans, with 25 spans sampled from each error category. Each span was evaluated for two criteria: whether it matched the intended category definition and whether its boundaries were plausible. Table~\ref{tab:silver_audit} reports the proportion of spans satisfying both criteria. This audit is intended as a quality check of the synthetic training data rather than as gold-standard validation.

\begin{table}[t]
\centering
\small
\setlength{\tabcolsep}{6pt}
\renewcommand{\arraystretch}{1.1}

\begin{tabular}{lccc}
\toprule
\textbf{Category} & \textbf{Audited} & \textbf{Valid} & \textbf{\% Valid} \\
\midrule
Coherence & 25 & 16 & 64.0 \\
Format & 25 & 21 & 84.0 \\
Lacks Synthesis & 25 & 21 & 84.0 \\
Unsupported Claim & 25 & 22 & 88.0 \\
\midrule
Overall & 100 & 80 & 80.0 \\
\bottomrule
\end{tabular}

\caption{Manual audit of critic-accepted synthetic spans. A span is counted as valid only if it satisfies both category validity and boundary plausibility.}
\label{tab:silver_audit}
\end{table}

\section{Model Evaluation Metrics}
\label{app:model_eval_metrics}

\paragraph{Pairwise F1}
Pairwise F1 is used to evaluate overlap between predicted and gold spans. The basic matching unit is a span, represented by token offsets \([s,e)\), where \(s\) is the start token index and \(e\) is the exclusive end token index. For each predicted span \(p\) and gold span \(g\) of the same category, we compute token-level Intersection-over-Union (IoU):

\begin{align}
    \mathrm{IoU}(p,g) =
    \frac{|T_p \cap T_g|}{|T_p \cup T_g|}
\end{align}

\noindent where \(T_p\) and \(T_g\) are the sets of token indices covered by the predicted and gold spans. Predicted and gold spans are matched one-to-one in descending IoU order. A predicted span is counted as a true positive if it matches a gold span with \(\mathrm{IoU} \geq 0.5\). Unmatched predicted spans are counted as false positives, and unmatched gold spans are counted as false negatives. Pairwise F1 is then computed as:

\begin{align}
    \mathrm{PairwiseF1} =
    \frac{2TP}{2TP + FP + FN}
\end{align}

\noindent Micro Pairwise F1 pools \(TP\), \(FP\), and \(FN\) across all evaluation documents before computing F1. Macro Pairwise F1 first computes Pairwise F1 separately for each document, then averages over documents. Thus, the micro score reflects overall span detection performance across the corpus, while the macro score reflects document-level consistency.

\paragraph{Cohen's $\kappa$}
Cohen's $\kappa$ is used as a stricter token-level agreement metric. Unlike Pairwise F1, which evaluates span overlap, Cohen's $\kappa$ treats each token as the evaluation unit. For each category, gold and predicted spans are converted into binary token sequences, where tokens inside a span are assigned label \(1\), and all other tokens are assigned label \(0\).

Let \(TP\), \(FP\), \(FN\), and \(TN\) denote the token-level confusion counts between the gold and predicted binary labels, and let

\begin{align}
    N = TP + FP + FN + TN .
\end{align}

\noindent The observed agreement is:

\begin{align}
    p_o = \frac{TP + TN}{N}.
\end{align}

\noindent The expected agreement by chance is computed from the marginal label distributions:

\begin{align}
    p_e =
    \left(\frac{TP + FN}{N}\right)
    \left(\frac{TP + FP}{N}\right)
    + \nonumber \\
    \left(\frac{FP + TN}{N}\right)
    \left(\frac{FN + TN}{N}\right)
\end{align}

\noindent Cohen's $\kappa$ is then defined as:

\begin{align}
    \kappa = \frac{p_o - p_e}{1 - p_e}.
\end{align}

\noindent Higher values indicate stronger token-level agreement beyond chance. In this task, Cohen's $\kappa$ is more sensitive to boundary mismatches than Pairwise F1, since extra or missing boundary tokens directly affect the token-level agreement calculation.

\section{Model Prediction Examples}
\label{app:model_prediction_examples}

\begin{table*}[t]
\centering
\scriptsize
\setlength{\tabcolsep}{4pt}
\renewcommand{\arraystretch}{1.15}

\begin{tabularx}{\textwidth}{
p{0.24\textwidth}
>{\raggedright\arraybackslash}X
>{\raggedright\arraybackslash}X
>{\raggedright\arraybackslash}X
>{\raggedright\arraybackslash}X
>{\raggedright\arraybackslash}X
}
\toprule
\textbf{In-text Example} 
& \textbf{ModernBERT}
& \textbf{SciBERT}
& \textbf{XLM-R}
& \textbf{NeoBERT}
& \textbf{Gold} \\
\midrule

``Following the work of DrQA (Chen et al., 2017), most recent works build a two-stage retriever-reader system to tackle the problem.''
& ``recent works build a two-stage retriever-reader system to tackle''
& ``recent works build a two-stage retriever-reader system to tackle''
& ``works build a two-stage retriever-reader system''
& \textemdash
& ``most recent works build a two-stage retriever-reader system to tackle the problem.'' \\

\midrule

``To verify the effectiveness of our method, we conduct extensive experiments on three large-scale benchmark datasets (OntoNotes V4.0, OntoNotes V5.0, and MSRA).''
& ``large-scale benchmark datasets''
& ``benchmark datasets''; ``V4.0, OntoNotes''
& ``V4.0, OntoNotes V5.0,''
& ``To verify the effectiveness of our method, we conduct extensive experiments on three large-scale benchmark datasets (OntoNotes V4.0, OntoNotes V5.0, and MSRA).''
& ``(OntoNotes V4.0,''; ``OntoNotes V5.0''; ``MSRA'' \\

\bottomrule
\end{tabularx}

\caption{Qualitative examples of model span predictions compared with gold annotations for {\em Unsupported Claim} spans.}
\label{tab:model_predictions}
\end{table*}

Table~\ref{tab:model_predictions} presents qualitative examples of model span predictions compared with gold annotations. These examples illustrate common prediction behaviors, including under-prediction, over-prediction, fragmented spans, and missed predictions. While several models identify the general region of the error, they often differ in how precisely they align with the gold span boundaries. This supports the quantitative results, where Pairwise F1 scores are higher than Cohen’s $\kappa$, indicating that models frequently capture approximate error regions without exact boundary agreement.

\section{Hyperparameter Values}
\label{app:hyperparameters}
We report the selected hyperparameter values used for the category-specific model configurations. Table~\ref{tab:hyperparameter_par} shows the selected values for the paragraph-level error categories, {\em Coherence} and {\em Lacks Synthesis}, including the coverage threshold \(t\), window and stride sizes in the sliding window classification setup. Table~\ref{tab:hyperparameter_sent} reports the hyperparameter values selected for BIO label weights for the token-level categories {\em Format} and {\em Long Unsupported Claim}, and Table~\ref{tab:hyperparameter_short} – the selected binary and singleton confidence thresholds used for {\em Short Unsupported Claim} span prediction. These values are included here for reproducibility.

\section{Synthetic Data \& Model Performance}
\label{app:synth_correlation}

To further examine the relationship between synthetic data quality and downstream performance, we compute Pearson correlations between distributional shifts in synthetic spans and model performance across error categories. The results are shown in Table~\ref{tab:synthetic_correlation}.

Higher token concentration, measured through Top-20 token coverage, shows a strong negative correlation with Cohen's $\kappa$, while lexical diversity metrics such as EAD-2 and vocabulary per span show positive correlations with $\kappa$. These trends suggest that synthetic spans with more repetitive lexical patterns are associated with weaker boundary agreement, while more lexically diverse synthetic spans are associated with stronger agreement.

Since these correlations are computed across a small number of error categories, they should be interpreted as indicative trends rather than definitive conclusions.

\begin{table}[t]
\centering
\small
\setlength{\tabcolsep}{6pt}
\renewcommand{\arraystretch}{1.1}

\begin{tabular}{lcc}
\toprule
\textbf{Metric} & \textbf{F1} & \textbf{$\kappa$} \\
\midrule
Top-20 Token Coverage & -0.395 & \textbf{-0.990} \\
EAD-2 (Lexical Diversity) & -0.257 & 0.779 \\
Vocabulary per Span & -0.470 & 0.612 \\
MATTR & -0.111 & 0.660 \\
Span Length & 0.325 & 0.498 \\
\bottomrule
\end{tabular}

\caption{Pearson correlation between synthetic data properties and model performance across categories.}
\label{tab:synthetic_correlation}
\end{table}

\section{Confidence Intervals}
\label{app:confidence_intervals}

To characterize uncertainty under the small and category-imbalanced evaluation set, we report 95\% bootstrap confidence intervals for Micro Pairwise F1 (see Tables \ref{tab:ci_format_micro}-\ref{tab:ci_lacks_synthesis_micro}). For each model and category, we resample evaluation units with replacement, recompute the score for each bootstrap sample, and report the 2.5th and 97.5th percentiles as the confidence interval. These intervals indicate the stability of the reported model scores; wider intervals suggest greater uncertainty, while overlapping intervals indicate that fine-grained model rankings should be interpreted cautiously.

The confidence intervals are reported for Micro Pairwise F1 because this is the primary span-overlap metric used in the main results. Pairwise F1 captures whether a model identifies the approximate region of a writing issue, whereas Cohen's $\kappa$ is used as a stricter measure of token-level boundary agreement. 

\begin{table}[t]
\centering
\small
\begin{tabular}{lcc}
\toprule
\textbf{Model} & \textbf{Micro Pairwise F1} & \textbf{95\% CI} \\
\midrule
XLM-R      & \underline{0.648} & $[0.570,\ 0.728]$ \\
SciBERT    & \textbf{0.667}    & $[0.583,\ 0.740]$ \\
ModernBERT & 0.533             & $[0.433,\ 0.634]$ \\
NeoBERT    & 0.053             & $[0.033,\ 0.075]$ \\
\bottomrule
\end{tabular}
\caption{Micro Pairwise F1 with 95\% bootstrap confidence intervals for
\textit{Format}.}
\label{tab:ci_format_micro}
\end{table}

\begin{table}[t]
\centering
\small
\begin{tabular}{lcc}
\toprule
\textbf{Model} & \textbf{Micro Pairwise F1} & \textbf{95\% CI} \\
\midrule
XLM-R      & \underline{0.263} & $[0.144,\ 0.383]$ \\
SciBERT    & \textbf{0.335}    & $[0.248,\ 0.423]$ \\
ModernBERT & 0.200             & $[0.104,\ 0.302]$ \\
NeoBERT    & 0.205             & $[0.123,\ 0.284]$ \\
\bottomrule
\end{tabular}
\caption{Micro Pairwise F1 with 95\% bootstrap confidence intervals for
\textit{Short Unsupported Claim}.}
\label{tab:ci_unsupported_short_micro}
\end{table}

\begin{table}[H]
\centering
\small
\begin{tabular}{lcc}
\toprule
\textbf{Model} & \textbf{Micro Pairwise F1} & \textbf{95\% CI} \\
\midrule
XLM-R      & \textbf{0.567}    & $[0.506,\ 0.626]$ \\
SciBERT    & 0.169             & $[0.108,\ 0.232]$ \\
ModernBERT & \underline{0.497} & $[0.437,\ 0.556]$ \\
NeoBERT    & 0.053             & $[0.024,\ 0.085]$ \\
\bottomrule
\end{tabular}
\caption{Micro Pairwise F1 with 95\% bootstrap confidence intervals for
\textit{Long Unsupported Claim}.}
\label{tab:ci_unsupported_long_micro}
\end{table}

\begin{table}[H]
\centering
\small
\begin{tabular}{lcc}
\toprule
\textbf{Model} & \textbf{Micro Pairwise F1} & \textbf{95\% CI} \\
\midrule
XLM-R      & 0.810             & $[0.754,\ 0.860]$ \\
SciBERT    & 0.975             & $[0.959,\ 0.989]$ \\
ModernBERT & \underline{0.985} & $[0.970,\ 0.997]$ \\
NeoBERT    & \textbf{0.988}    & $[0.974,\ 0.997]$ \\
\bottomrule
\end{tabular}
\caption{Micro Pairwise F1 with 95\% bootstrap confidence intervals for
\textit{Coherence}.}
\label{tab:ci_coherence_micro}
\end{table}

\begin{table}[H]
\centering
\small
\begin{tabular}{lcc}
\toprule
\textbf{Model} & \textbf{Micro Pairwise F1} & \textbf{95\% CI} \\
\midrule
XLM-R      & \textbf{0.990}    & $[0.980,\ 0.998]$ \\
SciBERT    & \underline{0.984} & $[0.972,\ 0.994]$ \\
ModernBERT & 0.708             & $[0.651,\ 0.771]$ \\
NeoBERT    & 0.816             & $[0.756,\ 0.861]$ \\
\bottomrule
\end{tabular}
\caption{Micro Pairwise F1 with 95\% bootstrap confidence intervals for
\textit{Lacks Synthesis}.}
\label{tab:ci_lacks_synthesis_micro}
\end{table}

\section{Prompts Used}
\label{sec:appx_prompts}
\begin{figure}[!htbp]
    \centering
    \includegraphics[width=0.9\linewidth]{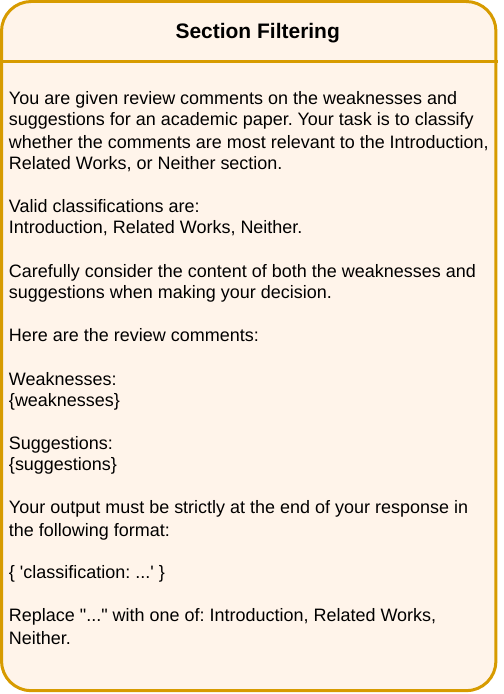}
    \caption{Prompt used to filter documents from the {\tt NLPeer} ARR subset based on whether reviewer comments refer to issues in the Introduction or Related Work sections.}
    \label{fig:filter_prompt}
\end{figure}

\begin{figure}[!htbp]
    \centering
    \includegraphics[width=0.95\linewidth]{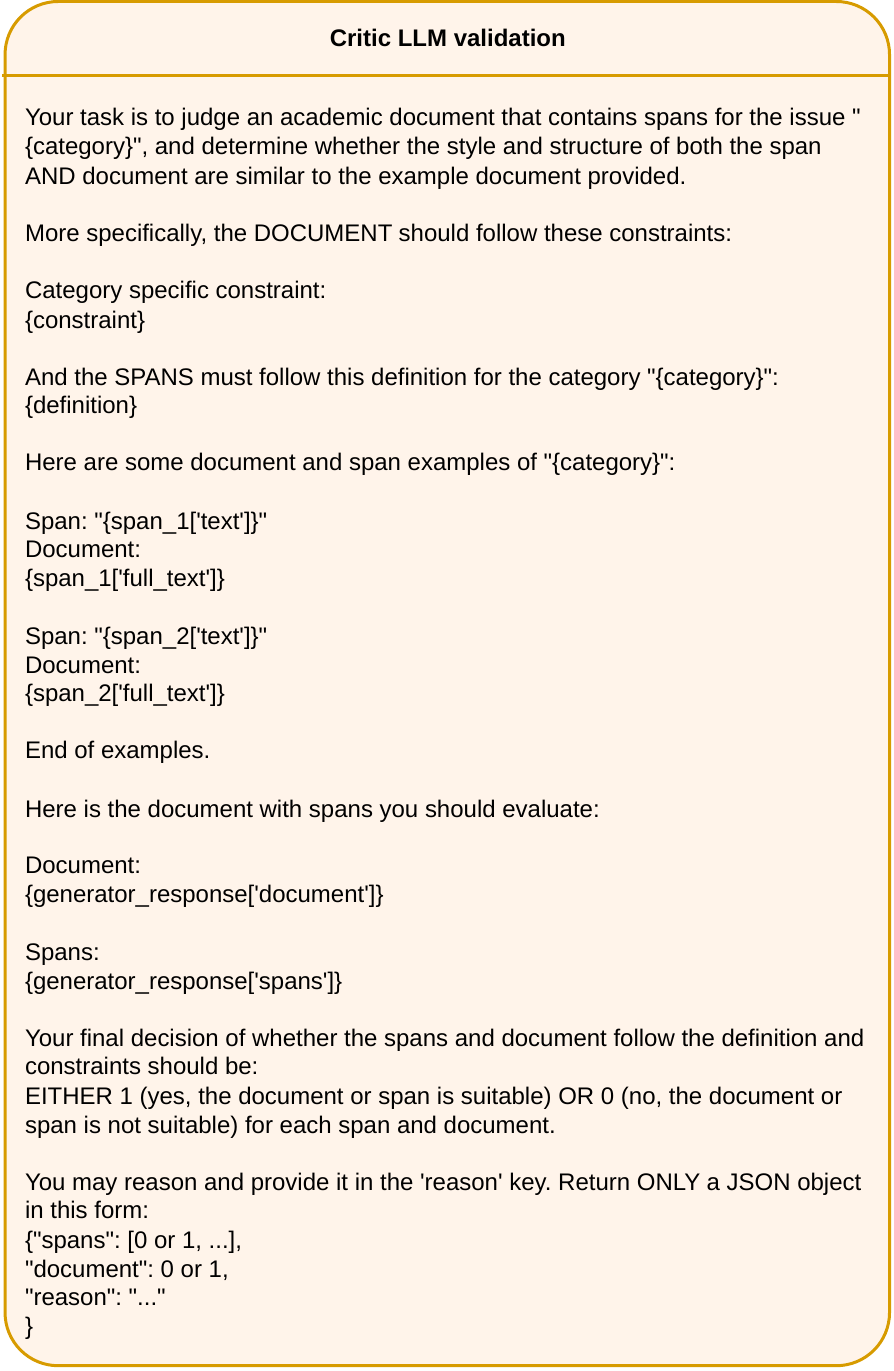}
    \caption{Critic prompt used to validate generated synthetic spans by checking whether each span matches the intended error category.}
    \label{fig:critic_prompt}
\end{figure}

\begin{figure}[!htbp]
    \centering
    \includegraphics[width=\linewidth]{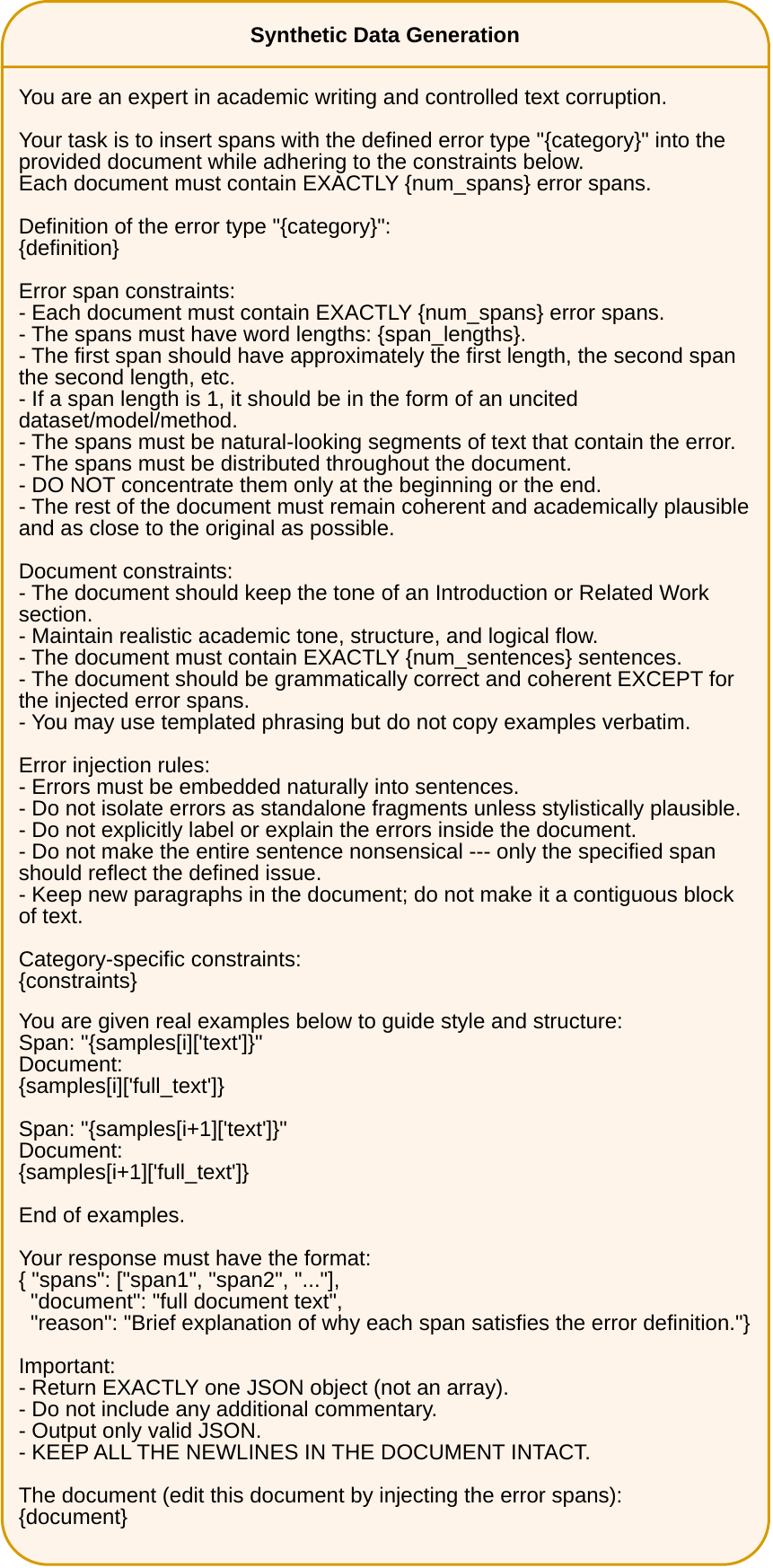}
    \caption{Prompt used to generate synthetic error spans from unused {\tt NLPeer} documents. The prompt includes category definitions, examples, and structural constraints sampled from the gold-standard annotations.}
    \label{fig:synth_generation_prompt}
\end{figure}

The prompts used for section filtering, synthetic span validation, and synthetic span generation are shown in Figures~\ref{fig:filter_prompt}, \ref{fig:critic_prompt}, and \ref{fig:synth_generation_prompt}.

\FloatBarrier

\end{document}